\documentclass[11pt]{article}

\usepackage[final]{acl}

\usepackage{times}
\usepackage{latexsym}
\usepackage[T1]{fontenc}
\usepackage[utf8]{inputenc}
\usepackage{microtype}
\usepackage{inconsolata}
\usepackage{graphicx}
\usepackage{booktabs}
\usepackage{multirow}
\usepackage{tabularx}
\usepackage{amsmath}
\usepackage{amssymb}
\usepackage{array}
\usepackage{enumitem}
\usepackage{tabularx}
\usepackage[most]{tcolorbox}
\tcbuselibrary{listings,breakable}
\usepackage{booktabs}
\usepackage{multicol}

\title{Value-Rich, Differently Organized: Measuring Value Expression and Organization in AI-Generated Debates}

\author{
  Maoyuan Li$^{1,2}$ \quad
  Haoyuan Li$^{2}$ \quad
  Zhongsheng Wang$^{2}$ \quad
  Jiamou Liu$^{2}$\thanks{Corresponding author.}
  \\
  $^{1}$Wuhan College of Communication \quad
  $^{2}$University of Auckland \\
  \texttt{mli302@aucklanduni.ac.nz},
  \texttt{hli962@aucklanduni.ac.nz} \\
  \texttt{zhongsheng.wang@auckland.ac.nz},
  \texttt{jiamou.liu@auckland.ac.nz}
}

\begin{document}
\maketitle

\begin{abstract}
Current evaluation of LLM-generated debate largely focuses on argumentative quality, leaving the value structure of generated arguments underexamined. We introduce a diagnostic metric framework for comparing how values are expressed and organized in human-reference and AI-generated debates. The framework separates \emph{value expression}, the range of recognizable values a debate invokes, from \emph{value organization}, how those values are distributed and sustained across turns. Grounded in Schwartz's Basic-10 value theory, it represents debates as multi-level value distributions over argumentative units, utterances, stances, and full debates, enabling metrics for value coverage, alignment, diffusion, and cross-turn dynamics. Across multiple generation systems, AI-generated debates are broadly value-expressive, often activating as many or more value dimensions than matched human references, but they organize these values differently. Their utterances are more locally value-diffuse, and source-blind judges distinguish value clarity from value focus. These findings show that debate evaluation should consider not only value presence, but also value organization. Code and supplementary materials are available at
\url{https://github.com/Maoyuan-li/Value-Expression-and-Organization}.
\end{abstract}

\section{Introduction}
\label{sec:introduction}

Arguments gain much of their force from the \emph{values} they bring into view. In debate, speakers do more than exchange claims and evidence: they reveal what they treat as fair, safe, free, dignified, or worth protecting. These value appeals shape why an argument feels compelling.

\begin{figure}[t]
    \centering
    \includegraphics[width=1\linewidth]{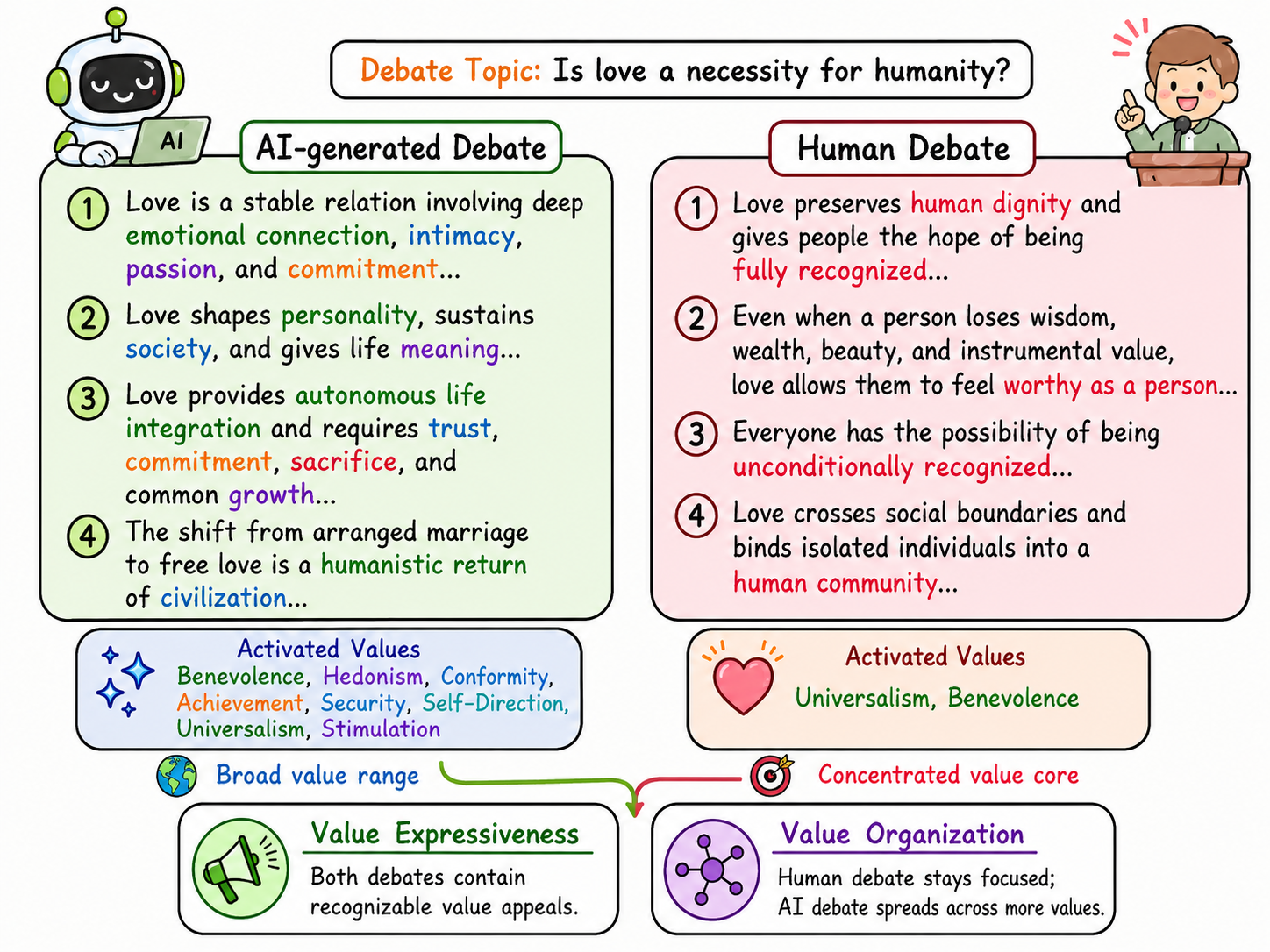}
    \caption{Illustrative contrast between value expression and value organization.}
    \label{fig:case-love}
\end{figure}

LLMs are increasingly able to generate fluent and persuasive debate utterances, and existing evaluation has largely focused on coherence, factuality, stance consistency, logical validity, and persuasiveness~\citep{liang2024debatrix,wang2025inspiredebate}. A separate line of value-oriented NLP has shown that human values can be identified in argumentative text, for example through Webis-ArgValues-22 and ValueEval~\citep{kiesel2022identifying,kiesel2023semeval}. Yet the intersection remains underexplored: debate evaluation rarely asks how values are organized across turns, while value mining usually treats values as labels attached to isolated texts. We therefore ask how AI-generated debates compare with matched human debate references in {\em value expression} and {\em value organization}. As Figure~\ref{fig:case-love} illustrates, a debate may be value-expressive by invoking recognizable value appeals, while differing in value organization: the AI-generated example spreads across a broader value range, whereas the human reference concentrates around a more compact core of Universalism and Benevolence.

We frame value-based debate analysis as comparing \emph{distributions of value appeals} across matched human-reference and AI-generated debates. This raises two challenges. First, values are rarely stated as explicit labels: a speaker may never name ``security'' or ``autonomy,'' while still embedding those values in the arguments. Second, values operate at multiple levels, from local argumentative units to utterances, stances, and full debates. We therefore adopt Schwartz's Basic-10 theory as a theory-grounded value space~\citep{schwartz1992universals,schwartz2012refining}. Following the claim--premise view of argument structure in \citet{stab2014annotating}, we represent each argumentative unit as a stance-specific claim paired with a premise span extracted from the corresponding utterance. Each argumentative unit is assigned a soft distribution over the ten values, which is then aggregated at multiple levels.

\begin{figure*}[t]
    \centering
    \includegraphics[width=1\textwidth]{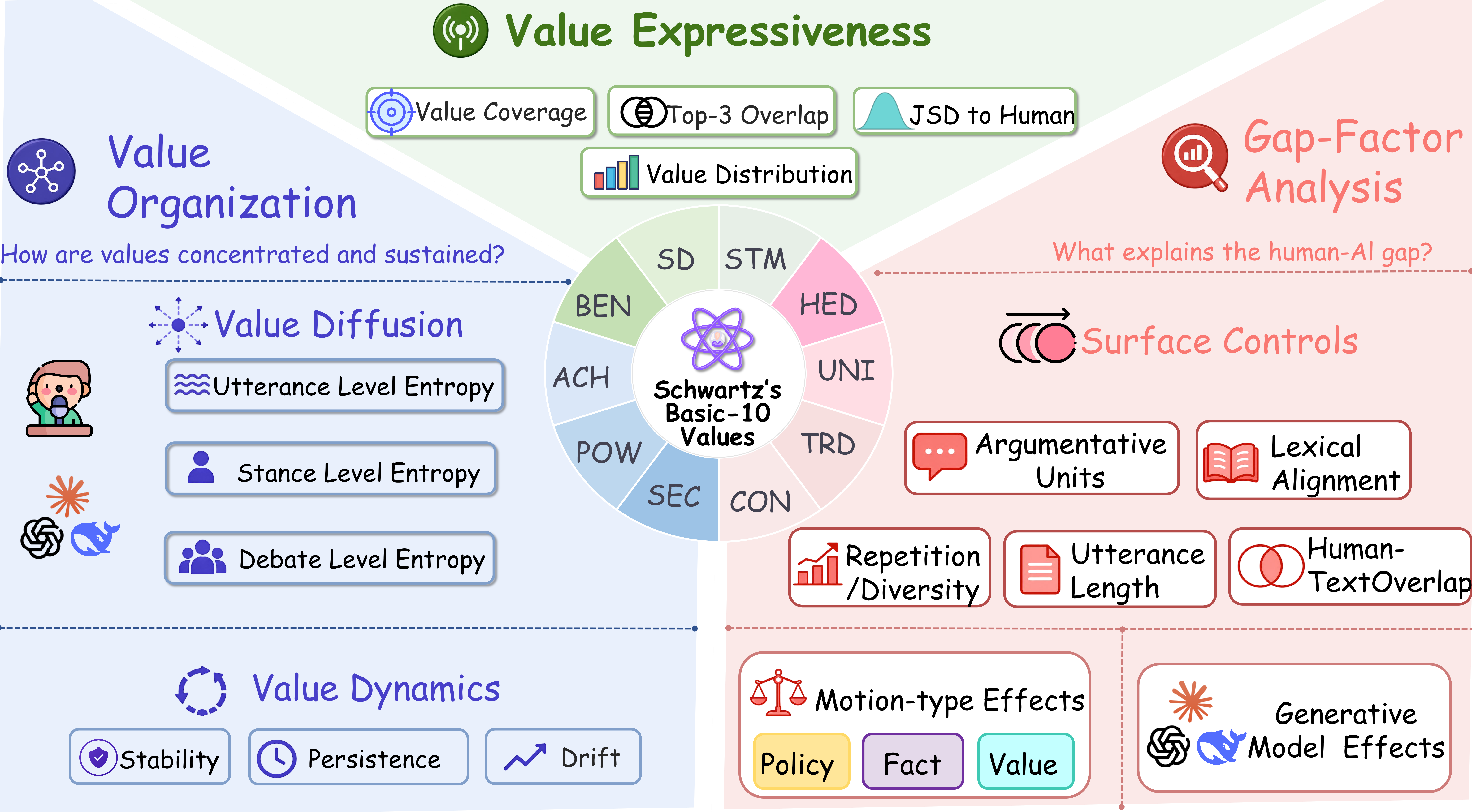}
    \caption{
  Conceptual framework. The shared Schwartz Basic-10 space (center) grounds two evaluation dimensions: value expression (breadth of value activation) and value organization (concentration and cross-turn dynamics). The right panel shows how paired differences are interpreted through surface controls, motion-type effects, and generation-system effects. Here, Human-TextOverlap denotes BLEU similarity between a generated debate and its matched human reference.}
    \label{fig:conceptual-framework}
\end{figure*}

This multi-level representation supports a diagnostic metric suite with two complementary components. For value expression, we measure how broadly a debate activates the Basic-10 value space and how closely its dominant values align with a paired human reference. For value organization, we define a family of metrics that treats debate values as both distributions and trajectories: entropy captures value diffusion, while stability, drift, and dominant-value persistence characterize how a stance carries its value profile across turns. These metrics shift attention from whether generated arguments mention human-recognizable values to how those values are structured and transformed within argumentative interaction.

We instantiate the framework in a paired benchmark comparing general-purpose LLMs with R-Debater (Li et al., 2025). Results show a consistent separation: AI-generated debates activate as many or more Basic-10 values as human references while distributing value mass differently across utterances and stances. These patterns persist after surface-level controls and are supported by source-blind human evaluation. 

Our contributions are: (1) a diagnostic distinction between value expression and value organization; (2) a Schwartz-grounded multi-level framework representing debate values as distributions over argumentative units, utterances, stances, and full debates; (3) a metric family covering value coverage, alignment, local concentration, and cross-turn dynamics; and (4) a paired analysis showing that AI-generated debates activate broad value spaces while differing from human references in local concentration and cross-turn organization. 

\section{Related Work}
\label{sec:related_work}

\noindent\textbf{Automated debate generation and evaluation.} Early automated debate studies focused on the components needed to build debating systems, such as evidence retrieval~\citep{rinott2015show}, argument construction~\citep{gutfreund2016automatic}, stance modeling~\citep{bar2017stance}, and argument quality assessment~\citep{bar2021advances,toledo2019automatic}. Project Debater later integrated retrieval, argument mining, and speech planning into an autonomous debating system~\citep{slonim2021autonomous}. Recent LLM-based work has shifted toward debate-like interaction, using simulated agents and multi-agent debate~\citep{du2024improving,liang2024encouraging}, retrieval-augmented debate generation~\citep{li2025rdebater}, and evaluation-guided optimization~\citep{wang2025inspiredebate}. Debate evaluation has also expanded from winning-stance prediction~\citep{ruiz2023automatic} to multi-dimensional judging, chronological analysis~\citep{liang2024debatrix,wang2025inspiredebate}, and LLM-as-judge benchmarking~\citep{sternlicht2025debatable}. 
However, none measures how values are distributed and sustained across debate turns.

\noindent\textbf{Human values in argumentation.} Values have long been central to argumentation theory. In value-based argumentation, argument acceptability depends not only on logical relations, but also on the values promoted by an argument and the audience's value ordering~\citep{benchcapon2003persuasion}. Early value-oriented argument mining focused on identifying implicit human values behind natural-language arguments~\citep{kiesel2022identifying}. Later work expanded this line with larger and more diverse resources, including Touch\'e23-ValueEval~\citep{mirzakhmedova2024touche23} and the SemEval-2023 ValueEval shared task~\citep{kiesel2023semeval}. Related value-modeling datasets also study human values in social scenarios, dialogue, or LLM outputs, such as ValueNet~\citep{qiu2022valuenet} and FULCRA~\citep{yao2024value}. 
However, these studies treat values as labels for isolated texts. We instead study them as distributions in matched multi-turn debates.


\section{Conceptual Framework}
\label{sec:conceptual-framework}

This section introduces the conceptual framework used to analyze value expression and value organization in human-reference and AI-generated debates. 
Figure~\ref{fig:conceptual-framework} summarizes the framework and its three layers: value representation, value-based measurement, and paired-difference interpretation.

\subsection{Shared Value Space}
\label{subsec:shared-value-space}

Value-based debate analysis requires a shared vocabulary for describing what values are invoked. We adopt Schwartz's Basic-10 theory as this value space~\citep{schwartz1992universals,schwartz2012refining}. 
Schwartz values are broad motivational goals guiding evaluations of actions and social arrangements, well suited to debate where arguments invoke safety, fairness, autonomy, or preservation. 
The Basic-10 level is broad enough to cover major motivational contrasts in debate. We use the framework to analyze textual value appeals expressed in debate discourse, not to infer the internal beliefs or moral commitments of a model. The same principle applies to human debate transcripts.
Table~\ref{tab:basic10-values} gives the acronyms used in Figure~\ref{fig:conceptual-framework} and the debate-oriented interpretation of each value. The ten values define the value set $\mathcal{V}$ used throughout the paper. Each value distribution in the framework is a probability distribution over this Basic-10 value space.


\begin{table}[t]
\centering
\footnotesize
\setlength{\tabcolsep}{4pt}
\renewcommand{\arraystretch}{0.85}
\begin{tabular}{p{0.13\linewidth}p{0.20\linewidth}p{0.55\linewidth}}
\toprule
\textbf{Acr.} & \textbf{Value} & \textbf{Debate-oriented interpretation} \\
\midrule
SD & Self-Direction & Autonomy, independent thought, freedom of choice. \\
STM & Stimulation & Novelty, challenge, innovation, change. \\
HED & Hedonism & Pleasure, comfort, enjoyment, quality of life. \\
ACH & Achievement & Success, competence, merit, productivity. \\
POW & Power & Authority, influence, status, control over resources. \\
SEC & Security & Safety, stability, public order, risk reduction. \\
CON & Conformity & Rule-following, responsibility, social norms. \\
TRD & Tradition & Heritage, customs, continuity, cultural preservation. \\
BEN & Benevolence & Care for close others, responsibility, loyalty. \\
UNI & Universalism & Equality, justice, human rights, social or environmental welfare. \\
\bottomrule
\end{tabular}
\caption{The Schwartz Basic-10 value space.}
\label{tab:basic10-values}
\end{table}

\subsection{Debate as a Multi-level Value Structure}
\label{subsec:multi-level-value-structure}

We represent a \emph{debate} formally as $d=(q,\mathcal{Z},\mathcal{C},\mathcal{U})$, where $q$ is the topic, $\mathcal{Z}=\{+,-\}$ denotes the pro and con stances, $\mathcal{C}=\{c_+,c_-\}$ contains the corresponding stance-specific claims, and $\mathcal{U}=(u_1,\ldots,u_n)$ is the ordered \emph{utterance} list. Each utterance $u_t$ has text $x_t$, stance $z_t\in\mathcal{Z}$, and turn index $t$. The two stances alternate in turn, i.e., $z_t \neq z_{t+1}$ for any $t$. 

Following the claim--premise view of argument structure in \citet{stab2014annotating}, we adapt this structure to competitive debate, where the stance-specific claim is determined by the debate topic and the speaker's stance. For an utterance $u_t$ with stance $z_t$, SAPE-Agent extracts premise spans $\mathcal{P}(u_t)=\{p_{t,1},\ldots,p_{t,k_t}\}$ from the utterance text $x_t$. We define an \emph{argumentative unit} as the claim--premise pair
\begin{equation}
    a_{t,i}=(c_{z_t},p_{t,i}),
\end{equation}
where $c_{z_t}$ is the stance-specific claim and $p_{t,i}$ is an extracted premise. Accordingly, $\mathcal{A}(u_t)=\{a_{t,1},\ldots,a_{t,k_t}\}$ denotes the argumentative units in utterance $u_t$. A premise may support the stance-specific claim directly or indirectly through rebuttal, counterexample, framing, concession, or value weighing, without requiring a fine-grained debate-act taxonomy. 
A Basic-10 value distribution is first assigned to each argumentative unit, then aggregated to three structural levels: utterance, stance, and debate. These levels correspond to increasingly broader views of value expression. Let $\mathbf{p}(a_{t,i})\in \Delta^{|\mathcal{V}|-1}$ denote the Basic-10 value distribution associated with argumentative unit $a_{t,i}$ in utterance $u_t$.
At the utterance level, we ask what value profile is expressed by one debate turn. For utterance $u_t$ with argumentative units $\mathcal{A}(u_t)=\{a_{t,1},\ldots,a_{t,k_t}\}$, the \emph{utterance distribution} is:\vspace*{-2mm}
 \begin{equation} \footnotesize
     \mathbf{P}_{\mathrm{utt}}(u_t)=
     \frac{1}{k_t}\sum_{a\in \mathcal{A}(u_t)}\mathbf{p}(a). \vspace*{-2mm}
 \end{equation}
Because it is an average of probability distributions, $\mathbf{P}_{\mathrm{utt}}(u_t)$ is also a probability distribution over $\mathcal{V}$.
At the stance level, we ask what value profile is carried by one stance across its same-stance utterances. Let $\mathcal{U}_z(d)=(u_{z,1},\ldots,u_{z,n_z})$ be the ordered subsequence of value-bearing utterances produced under stance $z$. The \emph{stance distribution} is
 \begin{equation}\footnotesize
     \mathbf{P}_{\mathrm{sta}}(d\mid z)=
     \frac{1}{n_z}\sum_{u\in \mathcal{U}_z(d)}\mathbf{P}_{\mathrm{utt}}(u).
 \end{equation}
At the debate level, we summarize the overall value profile of the full exchange. Let $\mathcal{U}^*(d)$ be the set of value-bearing utterances and $n^*=|\mathcal{U}^*(d)|$. The \emph{debate distribution} is
 \begin{equation}\footnotesize
     \mathbf{P}_{\mathrm{deb}}(d)=
     \frac{1}{n^*}\sum_{u_t\in\mathcal{U}^*(d)}\mathbf{P}_{\mathrm{utt}}(u_t).
 \end{equation}

\subsection{Evaluation Dimensions}
\label{subsec:evaluation-dimensions}

Table~\ref{tab:metric-summary} summarizes the metrics.


\begin{table*}[t]
\centering
\small
\setlength{\tabcolsep}{4pt}
\renewcommand{\arraystretch}{0.85}
\begin{tabular}{p{0.18\linewidth}p{0.22\linewidth}p{0.48\linewidth}}
\toprule
\textbf{Dimension} & \textbf{Metric} & \textbf{Conceptual question} \\
\midrule
Value Expression & Coverage & How many Basic-10 values are activated? \\
Value Expression & Top-$k$ overlap & How much do generated and reference debates share dominant values? \\
Value Expression & JSD & How close is a generated debate's value distribution to its matched reference? \\
\midrule
Value Organization & Entropy & How concentrated or diffuse are value distributions at utterance, stance, and debate levels? \\
Value Organization & Stability & How smoothly does a stance-level value profile change across turns? \\
Value Organization & Drift & How far does a stance move from its opening value profile? \\
Value Organization & Persistence & Do dominant values recur across adjacent same-stance utterances? \\
\bottomrule
\end{tabular}
\caption{Evaluation dimensions and metrics.}
\label{tab:metric-summary}
\end{table*}

\noindent\textbf{Value expression.} This dimension concerns the breadth and reference alignment of expressed values. It asks how broadly an AI-generated debate activates the Basic-10 value space and how its dominant values compare with those in the matched human reference.

\noindent \emph{1. Coverage} counts the number of value dimensions activated above a threshold $\tau$:
 \begin{equation}\footnotesize 
     \mathrm{Coverage}_{\tau}(d)=
     \sum_{v\in\mathcal{V}}
     \mathbb{I}\left[P_{\mathrm{deb}}(v\mid d)\geq \tau\right].
 \end{equation}

\smallskip

\noindent \emph{2. Top-$k$ overlap} measures agreement in dominant values. Let $\mathrm{Top}_k(d)$ denote the set of $k$ highest-scoring values in $\mathbf{P}_{\mathrm{deb}}(d)$. For a matched human reference $d^H$ and generated debate $d^M$,
 \begin{equation}\footnotesize
     \mathrm{Overlap}_k(d^H,d^M)=
     \frac{|\mathrm{Top}_k(d^H)\cap \mathrm{Top}_k(d^M)|}{k}.
 \end{equation}

\smallskip

\noindent \emph{3. Jensen--Shannon divergence} measures alignment between the full debate-level value distributions:
\begin{equation}\footnotesize
     \mathrm{JSD}\left(
     \mathbf{P}_{\mathrm{deb}}(d^H)
     \parallel
     \mathbf{P}_{\mathrm{deb}}(d^M)
     \right).
 \end{equation}
Lower JSD values correspond to closer distributional alignment in the Basic-10 value space. These metrics do not capture whether values are locally focused or consistently carried across the debate.

\noindent\textbf{Value organization.} This dimension concerns the internal shape of value expression. We measure both value concentration and value dynamics.

\smallskip

\noindent \emph{4. Value entropy} measures concentration. For any value distribution $\mathbf{P}$ over $\mathcal{V}$, define
 \begin{equation}\footnotesize
     H(\mathbf{P})=
     -\sum_{v\in\mathcal{V}}P(v)\log P(v).
 \end{equation}
Applied to $\mathbf{P}_{\mathrm{utt}}(u_t)$, $\mathbf{P}_{\mathrm{sta}}(d\mid z)$, and $\mathbf{P}_{\mathrm{deb}}(d)$, this gives $H_{\mathrm{utt}}(u_t)$, $H_{\mathrm{sta}}(d\mid z)$, and $H_{\mathrm{deb}}(d)$.
Value dynamics are computed within each stance, because a debate alternates between competing stances. These metrics ask how a stance carries its own value profile across same-stance utterances:

\smallskip

\noindent \emph{5. Value stability} measures how smoothly the utterance-level value distribution changes across same-stance utterances. Define $\mathrm{Stability}_z(d)$ as
\begin{equation}\footnotesize 
     1-\frac{1}{n_z-1}\sum_{i=1}^{n_z-1}
     \mathrm{JSD}\left(
     \mathbf{P}_{\mathrm{utt}}(u_{z,i})
     \parallel
     \mathbf{P}_{\mathrm{utt}}(u_{z,i+1})
     \right).
 \end{equation}

Higher stability means less abrupt value change. We report the debate-level stability as the average of the two stance-level stability scores.

\smallskip

\noindent \emph{6. Value drift} captures the opening-to-closing shift of a stance:
 \begin{equation}\footnotesize
     \mathrm{Drift}_z(d)=
     \mathrm{JSD}
     \bigl(
     \mathbf{P}_{\mathrm{utt}}(u_{z,1})
     \parallel
     \mathbf{P}_{\mathrm{utt}}(u_{z,n_z})
     \bigr).
\end{equation}

\smallskip

\noindent \emph{7. Dominant-value persistence} captures whether prominent values continue across adjacent same-stance utterances. Let $\mathrm{Top}_k(u)$ be the set of top-$k$ values in utterance $u$. We define $\mathrm{Persistence}_{k,z}(d)$ as
\begin{equation}\footnotesize 
     \frac{1}{n_z-1}
     \sum_{i=1}^{n_z-1}
     \mathbb{I}
     [\mathrm{Top}_k(u_{z,i})\cap \mathrm{Top}_k(u_{z,i+1})\neq\emptyset].
 \end{equation}
Together, these metrics distinguish broad value activation from structured value organization. 

\subsection{Interpreting Paired Human--AI Differences}
\label{subsec:interpreting-differences}
For any metric $\mu$, we compute the paired difference $\Delta\mu(i,g) = \mu(d^M_{i,g}) - \mu(d^H_i)$, where $d^H_i$ is the human reference for topic $i$ and $d^M_{i,g}$ is the generated debate from system $g$. Positive values indicate larger metric values for the generated debate; they do not imply higher quality. Later analyses ask whether these differences persist after controlling for surface form, motion type, and generation system.

\begin{figure*}[t]
    \centering
    \includegraphics[width=1\textwidth]{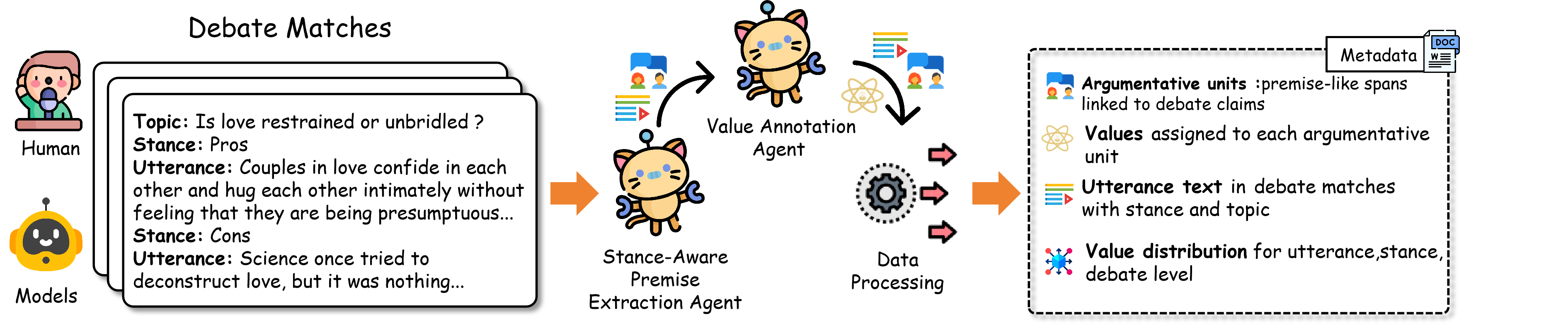}
    \caption{
    Operational pipeline. Transcripts are structured with metadata, processed by SAPE-Agent (premise extraction and argumentative-unit construction) and VA-Agent (Basic-10 value annotation), then aggregated into utterance-, stance-, and debate-level representations for metric computation. }
    \label{fig:pipeline}
\end{figure*}

\section{Methodology}
\label{sec:methodology}

Figure~\ref{fig:pipeline} illustrates the operational pipeline used to instantiate the conceptual framework. 
We structure transcripts, extract premise spans, construct argumentative units by pairing each premise with the corresponding stance-specific claim, assign Basic-10 value distributions, aggregate across levels, and compute the metrics in Section~\ref{sec:conceptual-framework}. Full prompts are provided in Appendix~\ref{app:prompts}.

\subsection{Transcript Structuring}
\label{subsec:transcript-structuring}

We convert each debate transcript into a structured record containing debate ID, system source, topic, motion type, stance, turn index, and utterance text. This structure aligns generated debates with their matched human references and determines the utterance sequences used for stance-level aggregation and same-stance cross-turn metrics.

\subsection{Premise Extraction and Argumentative-unit Construction}
\label{subsec:sape}

We implement a \emph{Stance-Aware Premise Extraction Agent} (SAPE-Agent) to identify premise spans in each utterance. Input: debate topic, stance-specific claims, utterance text, and local context. Output: verbatim premise spans; multiple premises may be extracted per utterance. Extraction is conditioned on the debate topic, stance, and stance-specific claims, so that spans are identified relative to the argumentative role they play in the debate. For an utterance $u_t$ with stance $z_t$, each extracted premise $p_{t,i}$ is paired with the corresponding stance-specific claim $c_{z_t}$ to construct the argumentative unit $a_{t,i}=(c_{z_t},p_{t,i})$. 

\subsection{Value Annotation}
\label{subsec:value-annotation}

We implement a \emph{Value Annotation Agent} (VA-Agent) to map each argumentative unit to soft scores over the Basic-10 value space $\mathcal{V}$. The input includes the argumentative unit, the debate topic, the stance, value definitions, and retrieved human-annotated examples when available. The output is a ten-dimensional Basic-10 value distribution.

Let $s_v(a)\in[0,1]$ be the average score assigned to value $v\in\mathcal{V}$ for argumentative unit $a$ over $R$ runs. The resulting vector $\mathbf{p}(a)=(p(v\mid a))_{v\in\mathcal{V}}$ is the Basic-10 value distribution used in subsequent aggregation, where
\begin{equation}
     p(v\mid a)=
     \frac{s_v(a)}{\sum_{v'\in\mathcal{V}}s_{v'}(a)}.
 \end{equation}

\subsection{Aggregation and Metric Computation}
\label{subsec:aggregation-computation}

After value annotation, we aggregate unit-level Basic-10 value distributions according to the three levels defined in Section~\ref{sec:conceptual-framework}: utterance, stance, and debate. Ordered same-stance utterances are retained for cross-turn organization metrics, including stability, drift, and dominant-value persistence. We then compute value expression and value organization metrics using the definitions in Section~\ref{sec:conceptual-framework}, together with the surface-control variables used in the empirical analysis.


We validate output format, field completeness, span boundaries, and value-vector dimensionality for all pipeline outputs. Malformed annotations are regenerated or removed, and repeated scoring with averaging reduces instability in value assignment. 

\section{Experimental Design}
\label{sec:experiments}

The experiments address three questions: (RQ1) whether SAPE-Agent and VA-Agent provide reliable annotations; (RQ2) how AI-generated debates compare with human references in value expression and organization; and (RQ3) whether paired differences persist after controlling for surface factors, motion type, and generation system.  RQ1 is addressed through annotation validation; RQ2–3 through paired comparison. Human evaluation is described in Section ~\ref{sec:human-evaluation}. 

\textbf{Annotation validation.}
We validate both agents on a manually checked subset. For SAPE-Agent, we measure observed agreement, Cohen's $\kappa$ \citep{cohen1960coefficient}, and Krippendorff's $\alpha$ \citep{krippendorff2004reliability} against human premise-span annotations.
For VA-Agent, human annotators assign Basic-10 value labels to sampled argumentative units. We evaluate VA-Agent as a multi-label value annotation model using precision, recall, and F1, and compare it with BERT, SVM, and direct LLM annotation baselines. The annotation protocol is shown in Appendix~\ref{app:annotation}.

\textbf{Benchmark construction.}
To compare human-reference and AI-generated debates under aligned conditions, we construct a benchmark from ORCHID~\citep{zhao2023orchid}, a Chinese text-only corpus of real-world competitive debates. ORCHID contains 1,218 Chinese debates across 476 unique topics, with 2,436 stance-specific summaries and 14,133 fully annotated utterances, making it suitable for evaluating argumentative dialogue in Chinese~\citep{zhao2023orchid}. For our study, we sample 60 distinct debate topics.

For each selected topic, we use one ORCHID debate as the human reference debate and generate corresponding AI debates using GPT-5.4, Claude-3.6 Sonnet, Qwen3.6-Plus, and DeepSeek-V3.2. We also include R-Debater~\citep{li2025rdebater} as a debate-specific generation baseline. 
Each topic is assigned one of three motion types: fact, value, or policy~\citep{rottenberg2011elements}, allowing us to examine whether paired differences vary with the argumentative demands associated with different motion types.

\begin{table}[t]
\centering
\small
\setlength{\tabcolsep}{4pt}
\renewcommand{\arraystretch}{0.85}
\begin{tabular}{p{0.33\linewidth}p{0.57\linewidth}}
\toprule
\textbf{Component} & \textbf{Description} \\
\midrule
Human source & ORCHID Chinese competitive debates \\
Topics & 60 sampled debate topics \\
Human reference & One human debate per topic \\
AI systems & GPT-5.4, Claude-3.6 Sonnet, Qwen3.6-Plus, DeepSeek-V3.2, R-Debater \\
Motion types & Fact, value, policy \\
Comparison & Paired by topic, stance setting, language, turn structure, and approximate utterance length \\
\bottomrule
\end{tabular}
\caption{Benchmark composition.}
\label{tab:benchmark-composition}
\end{table}

\textbf{Paired comparison and analysis.}
For each topic, the human reference debate is paired with AI-generated debates under the same topic, stance setting, language, turn structure, and approximate utterance length. This design reduces topic and format-level confounds and allows value metrics to be interpreted as within-topic paired differences. We compute the paired difference $\Delta_{\mu}(i,g)$ defined in Section~\ref{subsec:interpreting-differences}. Positive differences indicate larger metric values for the AI-generated debate relative to the matched human reference.
We report paired differences for all metrics in Table~\ref{tab:metric-summary}, using Wilcoxon signed-rank tests~\citep{wilcoxon1945individual}, Holm correction within each metric family~\citep{holm1979simple}, and Cohen's $d_z$ \citep{lakens2013calculating}. Statistical significance indicates systematic difference, not superior debate quality.
To test robustness, we regress paired differences on surface controls (token length, argumentative-unit count, lexical diversity, BLEU similarity, Self-BLEU repetition) with generation-system and motion-type fixed effects.
Generation-system effects capture differences among GPT-5.4, Claude-3.6 Sonnet, Qwen3.6-Plus, DeepSeek-V3.2, and R-Debater. Motion-type effects capture differences among fact, value, and policy motions.

\section{Results and Analysis}
\label{sec:results-analysis}


\subsection{Annotation Validation}
\label{subsec:results-annotation-validation}

The annotation results indicate that SAPE-Agent and VA-Agent provide reliable signals for aggregate value analysis. For premise extraction, SAPE-Agent shows strong agreement with human annotations: observed agreement reaches 0.8612, Cohen's $\kappa$ is 0.7124 at the micro level and 0.6897 at the macro level, and Krippendorff's $\alpha$ is 0.7035. The premise rate is also close between human annotation and SAPE-Agent, with 0.3840 for human annotation and 0.4075 for SAPE-Agent. Each identified premise is subsequently paired with the corresponding stance-specific claim to construct the argumentative units used in downstream value analysis. For Basic-10 value annotation, VA-Agent outperforms all baselines. It achieves precision 0.74, recall 0.91, and F$_1$ 0.82, compared with F$_1$ scores of 0.25 for BERT, 0.20 for SVM, and 0.64 for direct LLM annotation. The BERT, SVM, and direct LLM results are annotation baselines; they are not debate-generation systems in the paired benchmark. 

\begin{table}[t]
\centering
\small
\setlength{\tabcolsep}{5pt}
\renewcommand{\arraystretch}{0.75}
\begin{tabular}{lc}
\toprule
\multicolumn{2}{c}{\textbf{(a) SAPE-Agent validation}} \\
\midrule
\textbf{Metric} & \textbf{Value} \\
\midrule
Observed Agreement & 0.8612 \\
Cohen's $\kappa$ (micro) & 0.7124 \\
Cohen's $\kappa$ (macro) & 0.6897 \\
Krippendorff's $\alpha$ (nominal) & 0.7035 \\
Premise Rate (Human) & 0.3840 \\
Premise Rate (SAPE-Agent) & 0.4075 \\
\midrule
\multicolumn{2}{c}{\textbf{(b) VA-Agent validation}} \\
\midrule
\textbf{Model} & \textbf{P / R / F$_1$} \\
\midrule
BERT & 0.40 / 0.19 / 0.25 \\
SVM & 0.21 / 0.19 / 0.20 \\
LLM (GPT-4o) & 0.63 / 0.65 / 0.64 \\
VA-Agent & \textbf{0.74 / 0.91 / 0.82} \\
\bottomrule
\end{tabular}
\caption{
Annotation validation. SAPE-Agent is evaluated against human premise-span annotations; each identified premise is subsequently paired with the corresponding stance-specific claim to construct an argumentative unit. VA-Agent is evaluated on multi-label Basic-10 value prediction at the argumentative-unit level. 
}
\label{tab:annotation-validation}
\end{table}

\subsection{Paired Comparison of Value Expression}
\label{subsec:results-expressiveness}

\begin{figure}[t]
    \centering
    \includegraphics[width=0.8\linewidth]{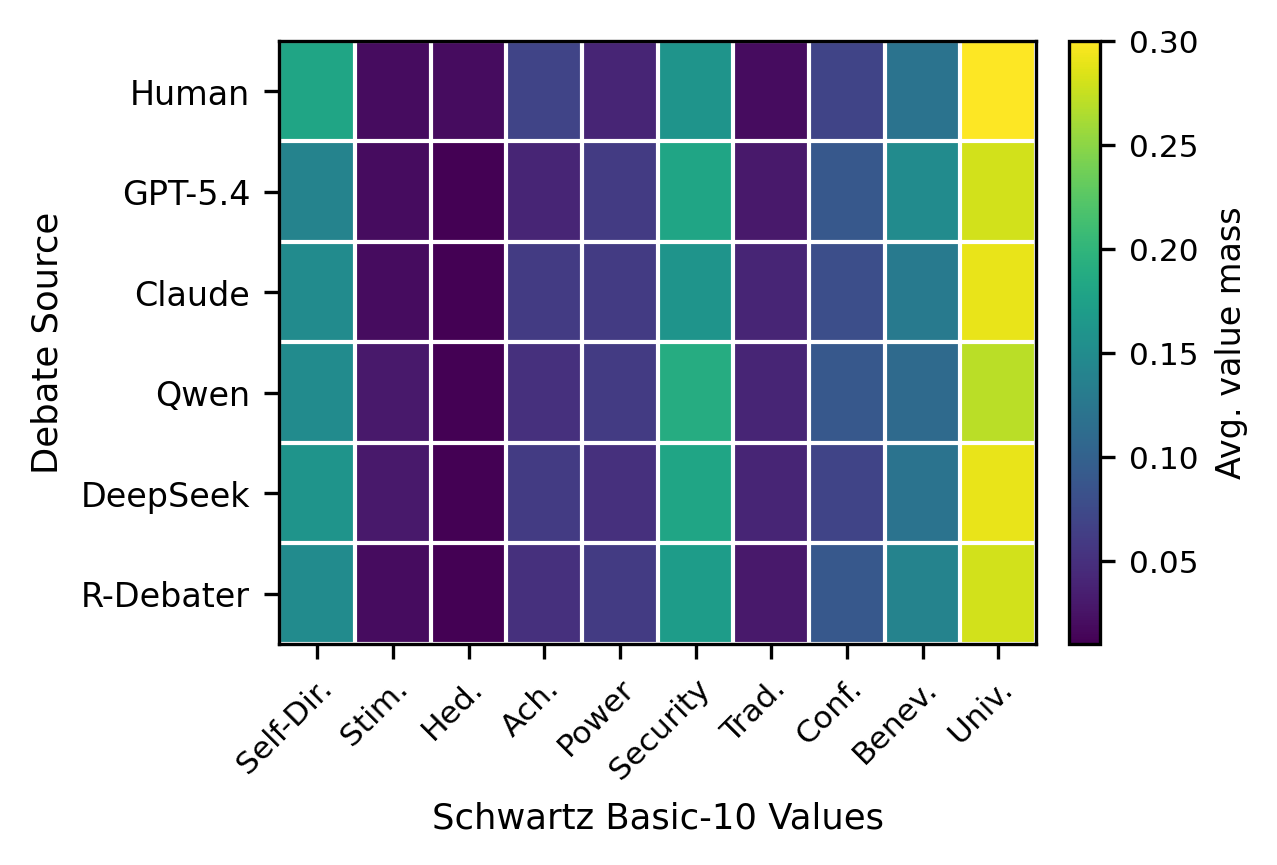}
    \caption{
    Aggregate Basic-10 value distributions across systems. Generation systems allocate value mass differently from matched human references. Color intensity indicates debate-level value mass. 
    }
    \label{fig:value-distribution-heatmap}
\end{figure}

AI-generated debates show broad value expression. Table~\ref{tab:value-expressiveness} summarizes the expressiveness metrics, and Figure~\ref{fig:value-distribution-heatmap} visualizes aggregate Basic-10 value distributions across human-reference and AI-generated debates. Human references activate 4.883 Basic-10 values on average, whereas AI-generated debates activate between 5.4 and 5.6 values. The largest coverage is observed for GPT-5.4 (5.600), followed by Qwen3.6-Plus (5.567) and R-Debater (5.520). Thus, generated debates in this benchmark do not show reduced value coverage; in many cases, they activate a broader range of value dimensions.

Broader coverage does not imply closer alignment with the matched human reference. Top-3 overlap ranges from 0.572 to 0.667, and JSD ranges from 0.102 to 0.133. DeepSeek-V3.2 has the closest distributional alignment, with the lowest JSD (0.102) and the highest Top-3 overlap (0.667), yet no system reproduces the reference value profile exactly. Together, Table~\ref{tab:value-expressiveness} and Figure~\ref{fig:value-distribution-heatmap} show that value expression and reference alignment are separable: generated debates can activate broad value spaces while allocating value mass differently from matched human debates.

\begin{table}[t]
\centering
\footnotesize
\setlength{\tabcolsep}{4pt}
\renewcommand{\arraystretch}{0.85}
\begin{tabular}{lccc}
\toprule
\textbf{System} & \textbf{Coverage} & \textbf{Top-3 Overlap} & \textbf{JSD} \\
\midrule
Human (reference)    & \textbf{4.883} & \textbf{1.000} & \textbf{0.000} \\
GPT-5.4   & 5.600 & 0.572 & 0.133 \\
Claude-3.6 Sonnet & 5.450 & 0.628 & 0.128 \\
Qwen3.6-Plus & 5.567 & 0.650 & 0.115 \\
DeepSeek-V3.2 & 5.400 & 0.667 & 0.102 \\
R-Debater & 5.520 & 0.633 & 0.123 \\
\bottomrule
\end{tabular}
\caption{Value expression metrics. Coverage denotes the average number of activated Basic-10 values using threshold $\tau=0.05$. Top-3 Overlap and JSD compare each system's debate-level value distribution with its matched human reference.}
\label{tab:value-expressiveness}
\end{table}

\subsection{Paired Comparison of Value Organization} \label{subsec:results-organization}

\begin{table}[!t]
\centering
\small
\setlength{\tabcolsep}{3.5pt}
\renewcommand{\arraystretch}{0.95}
\begin{tabularx}{\columnwidth}{@{}Xcccc@{}}
\toprule
\multicolumn{5}{c}{\textbf{(a) Utterance-level entropy $H_{\mathrm{utt}}$}} \\
\midrule
\textbf{System} & \textbf{Mean entropy} & \textbf{$\Delta H$} & \textbf{Holm $p$} & \textbf{$d_z$} \\
\midrule
Human (ref)     & 1.252 & --     & --        & -- \\
GPT-5.4   & 1.470 & +0.218 & $3.45{\times}10^{-7}$ & 0.848 \\
Claude-3.6 Sonnet & 1.476 & +0.224 & $3.45{\times}10^{-7}$ & 0.901 \\
Qwen3.6-Plus & 1.517 & +0.265 & $2.93{\times}10^{-8}$ & 1.018 \\
DeepSeek-V3.2 & 1.553 & +0.301 & $1.98{\times}10^{-9}$ & 1.301 \\
R-Debater & 1.532 & +0.280 & $3.71{\times}10^{-7}$ & 0.971 \\
\midrule
\multicolumn{5}{c}{\textbf{(b) Stance-level entropy $H_{\mathrm{sta}}$}} \\
\midrule
\textbf{System} & \textbf{Mean entropy} & \textbf{$\Delta H$} & \textbf{Holm $p$} & \textbf{$d_z$} \\
\midrule
Human (ref)     & 1.541 & --     & --        & -- \\
GPT-5.4   & 1.712 & +0.171 & $7.63{\times}10^{-5}$ & 0.637 \\
Claude-3.6 Sonnet & 1.756 & +0.215 & $1.40{\times}10^{-6}$ & 0.828 \\
Qwen3.6-Plus & 1.592 & +0.051 & $4.12{\times}10^{-3}$ & 0.397 \\
DeepSeek-V3.2 & 1.634 & +0.092 & $4.39{\times}10^{-3}$ & 0.409 \\
R-Debater & 1.649 & +0.108 & $8.40{\times}10^{-5}$ & 0.669 \\
\midrule
\multicolumn{5}{c}{\textbf{(c) Debate-level entropy $H_{\mathrm{deb}}$}} \\
\midrule
\textbf{System} & \textbf{Mean entropy} & \textbf{$\Delta H$} & \textbf{Holm $p$} & \textbf{$d_z$} \\
\midrule
Human (ref)    & 1.651 & --     & --      & -- \\
GPT-5.4   & 1.712 & +0.062 & 0.139 & 0.228 \\
Claude-3.6 Sonnet & 1.756 & +0.105 & 0.012  & 0.403 \\
Qwen3.6-Plus & 1.722 & +0.072 & 0.033 & 0.269 \\
DeepSeek-V3.2 & 1.744 & +0.093 & 0.016 & 0.382 \\
R-Debater & 1.761 & +0.110 & 0.026 & 0.323 \\
\bottomrule
\end{tabularx}
\caption{Multi-level value entropy. $\Delta H$ is computed as AI-generated debate minus its matched human reference at the corresponding aggregation level. Holm $p$ denotes the Wilcoxon signed-rank $p$-value after Holm correction within each level, and $d_z$ is the paired standardized effect size. Higher entropy indicates a more diffuse value distribution and is interpreted descriptively rather than as a direct quality score. All paired differences are significant after Holm correction except GPT-5.4 at the debate level.}
\label{tab:multi-level-entropy}
\end{table}

The strongest paired difference appears at the utterance level. Table~\ref{tab:multi-level-entropy} reports value entropy at all three aggregation levels. Human-reference utterances have a mean entropy of 1.252, while AI-generated utterances range from 1.470 to 1.553. DeepSeek-V3.2 shows the largest utterance-level paired difference ($\Delta H_{\mathrm{utt}}=+0.301$, $d_z=1.301$), followed by Qwen3.6-Plus ($\Delta H_{\mathrm{utt}}=+0.265$, $d_z=1.018$). The other systems also show large utterance-level differences, with $d_z$ between 0.848 and 0.971. This indicates that generated utterances tend to distribute value mass across a broader range of Basic-10 dimensions.

The paired difference becomes smaller as value distributions are aggregated. At the stance level, human-reference entropy is 1.541, while AI-generated debates range from 1.592 to 1.756. Claude-3.6 Sonnet shows the largest stance-level paired difference ($\Delta H_{\mathrm{sta}}=+0.215$, $d_z=0.828$), followed by GPT-5.4 ($\Delta H_{\mathrm{sta}}=+0.171$, $d_z=0.637$). At the debate level, the pattern is weaker and more generation-system dependent: human-reference debate-level entropy is 1.651, while AI-generated debates range from 1.712 to 1.761. GPT-5.4 has a small and non-significant debate-level difference ($\Delta H_{\mathrm{deb}}=+0.062$, Holm $p=0.139$, $d_z=0.228$). This suggests that the main value-organization difference is local: full-debate aggregation can partially mask the value diffusion visible in individual utterances.

Cross-turn dynamics show a second pattern. Human references have a stability score of 0.831, while AI-generated debates are higher, ranging from 0.915 for Claude-3.6 Sonnet to 0.952 for Qwen3.6-Plus. Qwen3.6-Plus, DeepSeek-V3.2, and R-Debater also have low drift values (0.047, 0.051, and 0.078) and perfect dominant-value persistence (1.000), while human references show higher drift (0.213) and slightly lower persistence (0.983). These results indicate smoother cross-turn value trajectories for several generation systems. Near-perfect persistence should be interpreted descriptively: it may reflect coherent value maintenance, but it may also indicate reduced adaptive movement across the exchange.


\begin{table}[t]
\centering
\small
\setlength{\tabcolsep}{7pt}
\begin{tabular}{lccc}
\toprule
\textbf{System} & \textbf{Stability} & \textbf{Drift} & \textbf{Persistence} \\
\midrule
Human     & 0.831 & 0.213 & 0.983 \\
GPT-5.4   & 0.933 & 0.217 & 0.988 \\
Claude-3.6 Sonnet & 0.915 & 0.221 & 0.979 \\
Qwen3.6-Plus & 0.952 & 0.047 & 1.000 \\
DeepSeek-V3.2 & 0.950 & 0.051 & 1.000 \\
R-Debater & 0.928 & 0.078 & 1.000 \\
\bottomrule
\end{tabular}
\caption{Cross-turn value organization. Stability measures how smoothly utterance-level value distributions change across same-stance utterances. Drift measures the opening-to-closing change in a stance's value distribution. Persistence measures whether dominant values are repeatedly maintained across adjacent same-stance utterances and persistence is computed with k = 3. Higher stability or persistence is not necessarily better; near-perfect persistence may indicate very smooth value maintenance, but should not be read as inherently superior.}
\label{tab:value-organization}
\end{table}

\subsection{Analysis of Paired Human--AI Differences}
\label{subsec:results-paired-differences}

We ask whether the paired differences can be explained by surface textual factors. Tab.~\ref{tab:surface-controls} shows cross-system variation in length, argumentative-unit count, lexical diversity, BLEU similarity, and Self-BLEU; Tab.~\ref{tab:surface-regression} controls for these factors. The adjusted differences remain positive for all dependent variables, with the strongest and most robust effect at the utterance level ($\Delta H_{\mathrm{utt}}=+0.252$, 95\% CI $[+0.205,+0.300]$). This suggests local value diffusion is not reducible to surface form.

Generation-system effects are significant only for stability ($p=.003$), indicating clearer system-level variation in cross-turn dynamics than in entropy. Motion-type effects are significant for all dependent variables, and Tab.~\ref{tab:motion-type-analysis} shows: fact motions have the largest debate-level entropy increase, value motions are more model-dependent, and policy motions are heterogeneous. Thus, paired differences persist after surface controls, but their magnitude and direction depend on motion type.

\section{Human Evaluation}
\label{sec:human-evaluation}


\textbf{Setup.}
We conduct a source-blind evaluation with 120 judges from university debate teams. The evaluation contains two complementary judgment components. Judges rate debates on overall quality, value clarity, value focus, perceived value stability, and perceived human-debate likeness using 7-point Likert scales. They also perform source-blind pairwise comparisons between debates on the same topic drawn from different sources and choose which debate better maintains a focused value core. Debate sources are hidden and the presentation order is randomized.

\begin{figure}[t]
    \centering
    \includegraphics[width=1\linewidth]{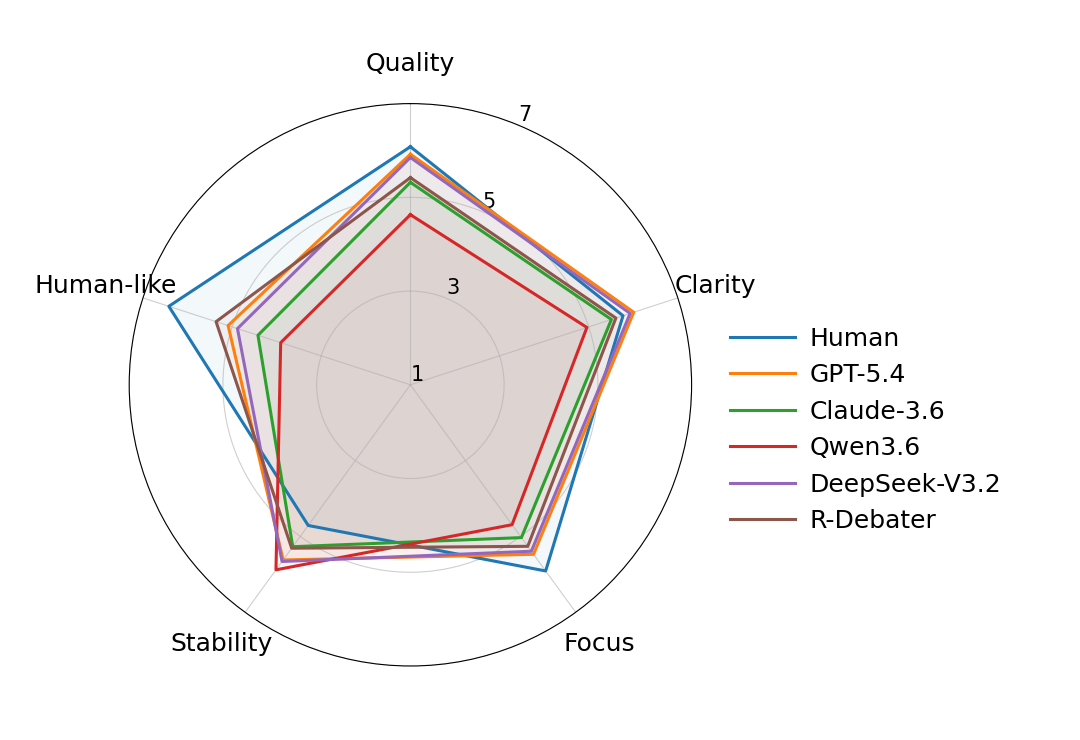}
    \caption{
    Source-blind human evaluation on five dimensions.}
    \label{fig:human-eval-radar}
\end{figure}

\textbf{Results and analysis.}
Figure~\ref{fig:human-eval-radar} and Table~\ref{tab:human-evaluation} show a clear separation between value clarity and value focus. Human references receive the highest perceived human-debate likeness score ($6.42 \pm 0.38$), while the closest AI-generated debates are GPT-5.4 ($5.09 \pm 0.47$) and DeepSeek-V3.2 ($4.88 \pm 0.49$). By contrast, AI-generated debates receive high value-clarity ratings: GPT-5.4 obtains the highest clarity score ($6.01 \pm 0.46$), followed by DeepSeek-V3.2 ($5.93 \pm 0.49$), both above the human-reference mean ($5.77 \pm 0.61$). The contrast is most visible in value focus. Human references receive the highest focus score ($5.91 \pm 0.56$) and are selected in 74\% of the pairwise focus comparisons involving a human-reference debate. These results support the paper's central distinction: generated debates can present value appeals clearly while differing from human references in how those appeals are focused and organized.

\section{Qualitative Case Study}
\label{sec:case-study}

Table~\ref{tab:case-study-love} compares human-reference and R-Debater utterances on "Love is a necessity for humanity." Human turns concentrate on Universalism and Benevolence; R-Debater activates care, pleasure, conformity, achievement, security, autonomy, and stimulation within the same motion—illustrating local value diffusion without implying either pattern is superior.  The case makes the entropy result interpretable: the difference lies in how selectively and consistently values are organized within a local argumentative frame.

\section{Conclusion}
\label{sec:conclusion}

We introduced a diagnostic framework for measuring value expression and value organization in AI-generated debate. Our results show that generated debates often invoke broad human-recognizable values, while differing from human references in local concentration and cross-turn dynamics. Debate evaluation should therefore consider not only which values appear, but how value appeals are structured across argumentative exchange.

\section*{Limitations}
\label{sec:limitations}

Our framework studies textual value appeals through Schwartz's Basic-10 value space. This provides a compact and theory-grounded basis for comparing human-reference and AI-generated debates, but it cannot cover every culturally specific, ideological, religious, rhetorical, or domain-specific value that may appear in debate. Other value frameworks may reveal complementary patterns.

Our main benchmark is based on Chinese competitive debate transcripts and a fixed-turn matched evaluation design. This improves comparability across human-reference and AI-generated debates, but it simplifies real debate interaction, where speakers may respond to interruptions, audience feedback, identity cues, and spontaneous strategic shifts. The cross-lingual analysis in Appendix~\ref{app:crosslingual} is exploratory, and broader multilingual evaluation remains future work.

The analysis also depends on automatic premise extraction, subsequent argumentative-unit construction, and Basic-10 value annotation. Although SAPE-Agent and VA-Agent are validated against human annotations, errors may remain for implicit, metaphorical, or culturally specific value appeals. In addition, value entropy, stability, drift, and persistence are descriptive measures of value organization, not direct quality scores. High entropy may indicate value diffusion, but it may also reflect inclusive or multi-perspective argumentation. For this reason, the framework is best understood as a lens for comparison and diagnosis. It identifies systematic organizational patterns, but it does not by itself determine whether a given value pattern is rhetorically desirable, ethically preferable, or strategically effective.

Finally, our results concern values expressed in generated debate text. They should not be interpreted as evidence about a model's internal beliefs, moral commitments, or alignment state. Nor do they replace evaluations of factuality, persuasiveness, or debate success. The contribution is instead to show that value organization is a distinct and necessary dimension of debate evaluation.

\section*{Ethical Considerations}

This work studies values expressed in debate text. We do not infer private moral commitments of individual speakers, demographic groups, or model users, nor do we treat generated value appeals as evidence of a model's internal beliefs or moral alignment state. Our analysis is conducted at the level of debate utterances and aggregate human-reference/AI-generation comparisons.

The human evaluation is source-blind and asks judges to assess debate texts. We collected limited demographic and debate-background information, including age, gender, education, and debate experience, solely for sample characterization and report these data only in aggregate. The main risks are annotator burden and possible exposure to controversial debate topics; these should be mitigated through clear task descriptions, informed consent, fair compensation, and the option to withdraw.

A value-based debate evaluation framework could be misused to optimize persuasive messaging or target audiences through value appeals. Our intended use is different: to provide diagnostic tools for evaluating, auditing, and understanding AI-generated debate. We therefore frame the metrics as descriptive measures of value expression and organization, not as tools for manipulating audiences, ranking speakers morally, or declaring one debate source inherently superior.

The ORCHID corpus is used in accordance with its publicly available release terms. Code and supplementary materials for our pipeline and paired benchmark are publicly available at \url{https://github.com/Maoyuan-li/Value-Expression-and-Organization}.

\bibliography{new}

@inproceedings{liang2024debatrix,
  title = {{Debatrix}: Multi-dimensional Debate Judge with Iterative Chronological Analysis Based on {LLM}},
  author = {Liang, Jingcong and Ye, Rong and Han, Meng and Lai, Ruofei and Zhang, Xinyu and Huang, Xuanjing and Wei, Zhongyu},
  booktitle = {Findings of the Association for Computational Linguistics: ACL 2024},
  pages = {14575--14595},
  year = {2024},
  address = {Bangkok, Thailand},
  publisher = {Association for Computational Linguistics},
  doi = {10.18653/v1/2024.findings-acl.868},
  url = {https://aclanthology.org/2024.findings-acl.868/}
}

@inproceedings{wang2025inspiredebate,
  title = {{InspireDebate}: Multi-Dimensional Subjective-Objective Evaluation-Guided Reasoning and Optimization for Debating},
  author = {Wang, Fuyu and Li, Jiangtong and Zhu, Kun and Jiang, Changjun},
  booktitle = {Proceedings of the 63rd Annual Meeting of the Association for Computational Linguistics (Volume 1: Long Papers)},
  pages = {27525--27544},
  year = {2025},
  address = {Vienna, Austria},
  publisher = {Association for Computational Linguistics},
  doi = {10.18653/v1/2025.acl-long.1335},
  url = {https://aclanthology.org/2025.acl-long.1335/}
}

@inproceedings{kiesel2022identifying,
  title = {Identifying the Human Values behind Arguments},
  author = {Kiesel, Johannes and Alshomary, Milad and Handke, Nicolas and Cai, Xiaoni and Wachsmuth, Henning and Stein, Benno},
  booktitle = {Proceedings of the 60th Annual Meeting of the Association for Computational Linguistics (Volume 1: Long Papers)},
  pages = {4459--4471},
  year = {2022},
  address = {Dublin, Ireland},
  publisher = {Association for Computational Linguistics},
  doi = {10.18653/v1/2022.acl-long.306},
  url = {https://aclanthology.org/2022.acl-long.306/}
}

@inproceedings{kiesel2023semeval,
  title = {{SemEval}-2023 Task 4: {ValueEval}: Identification of Human Values behind Arguments},
  author = {Kiesel, Johannes and Alshomary, Milad and Mirzakhmedova, Nailia and Heinrich, Maximilian and Handke, Nicolas and Wachsmuth, Henning and Stein, Benno},
  booktitle = {Proceedings of the 17th International Workshop on Semantic Evaluation (SemEval-2023)},
  pages = {2287--2303},
  year = {2023},
  address = {Toronto, Canada},
  publisher = {Association for Computational Linguistics},
  doi = {10.18653/v1/2023.semeval-1.313},
  url = {https://aclanthology.org/2023.semeval-1.313/}
}

@incollection{schwartz1992universals,
  title = {Universals in the Content and Structure of Values: Theoretical Advances and Empirical Tests in 20 Countries},
  author = {Schwartz, Shalom H.},
  booktitle = {Advances in Experimental Social Psychology},
  editor = {Zanna, Mark P.},
  volume = {25},
  pages = {1--65},
  year = {1992},
  publisher = {Academic Press},
  doi = {10.1016/S0065-2601(08)60281-6}
}

@article{schwartz2012refining,
  title = {Refining the Theory of Basic Individual Values},
  author = {Schwartz, Shalom H. and Cieciuch, Jan and Vecchione, Michele and Davidov, Eldad and Fischer, Ronald and Beierlein, Constanze and Ramos, Alice and Verkasalo, Markku and L{\"o}nnqvist, Jan-Erik and Demirutku, Kursad and Dirilen-Gumus, Ozlem and Konty, Mark},
  journal = {Journal of Personality and Social Psychology},
  volume = {103},
  number = {4},
  pages = {663--688},
  year = {2012},
  publisher = {American Psychological Association},
  doi = {10.1037/a0029393}
}

@inproceedings{stab2014annotating,
  title = {Annotating Argument Components and Relations in Persuasive Essays},
  author = {Stab, Christian and Gurevych, Iryna},
  booktitle = {Proceedings of {COLING} 2014, the 25th International Conference on Computational Linguistics: Technical Papers},
  pages = {1501--1510},
  year = {2014},
  address = {Dublin, Ireland},
  publisher = {Dublin City University and Association for Computational Linguistics},
  url = {https://aclanthology.org/C14-1142/}
}

@article{li2025rdebater,
  title = {{R-Debater}: Retrieval-Augmented Debate Generation through Argumentative Memory},
  author = {Li, Maoyuan and Wang, Zhongsheng and Li, Haoyuan and Liu, Jiamou},
  journal = {arXiv preprint arXiv:2512.24684},
  year = {2025},
  eprint = {2512.24684},
  archivePrefix = {arXiv},
  primaryClass = {cs.CL},
  url = {https://arxiv.org/abs/2512.24684}
}

@inproceedings{rinott2015show,
  title = {Show Me Your Evidence: An Automatic Method for Context Dependent Evidence Detection},
  author = {Rinott, Ruty and Dankin, Lena and Alzate Perez, Carlos and Khapra, Mitesh M. and Aharoni, Ehud and Slonim, Noam},
  booktitle = {Proceedings of the 2015 Conference on Empirical Methods in Natural Language Processing},
  pages = {440--450},
  year = {2015},
  address = {Lisbon, Portugal},
  publisher = {Association for Computational Linguistics},
  doi = {10.18653/v1/D15-1050},
  url = {https://aclanthology.org/D15-1050/}
}

@inproceedings{gutfreund2016automatic,
  title = {Automatic Arguments Construction: From Search Engine to Research Engine},
  author = {Gutfreund, Dan and Katz, Yoav and Slonim, Noam},
  booktitle = {{AAAI} Fall Symposium Series},
  year = {2016},
  url = {https://aaai.org/papers/14100-14100-automatic-arguments-construction-from-search-engine-to-research-engine/}
}

@inproceedings{bar2017stance,
  title = {Stance Classification of Context-Dependent Claims},
  author = {Bar-Haim, Roy and Bhattacharya, Indrajit and Dinuzzo, Francesco and Saha, Amrita and Slonim, Noam},
  booktitle = {Proceedings of the 15th Conference of the European Chapter of the Association for Computational Linguistics: Volume 1, Long Papers},
  pages = {251--261},
  year = {2017},
  address = {Valencia, Spain},
  publisher = {Association for Computational Linguistics},
  doi = {10.18653/v1/E17-1024},
  url = {https://aclanthology.org/E17-1024/}
}

@inproceedings{bar2021advances,
  title={Advances in debating technologies: Building AI that can debate humans},
  author={Bar-Haim, Roy and Dor, Liat Ein and Orbach, Matan and Venezian, Elad and Slonim, Noam},
  booktitle={Proceedings of the 59th Annual Meeting of the Association for Computational Linguistics and the 11th International Joint Conference on Natural Language Processing: Tutorial Abstracts},
  pages={1--5},
  year={2021}
}

@inproceedings{toledo2019automatic,
  title={Automatic argument quality assessment-new datasets and methods},
  author={Toledo, Assaf and Gretz, Shai and Cohen-Karlik, Edo and Friedman, Roni and Venezian, Elad and Lahav, Dan and Jacovi, Michal and Aharonov, Ranit and Slonim, Noam},
  booktitle={Proceedings of the 2019 Conference on Empirical Methods in Natural Language Processing and the 9th International Joint Conference on Natural Language Processing (EMNLP-IJCNLP)},
  pages={5625--5635},
  year={2019}
}

@article{slonim2021autonomous,
  title={An autonomous debating system},
  author={
    Slonim, Noam and
    Bilu, Yonatan and
    Alzate, Carlos and
    Bar-Haim, Roy and
    Bogin, Ben and
    Bonin, Francesca and
    Choshen, Leshem and
    Cohen-Karlik, Edo and
    Dankin, Lena and
    Edelstein, Lilach and
    Ein-Dor, Liat and
    Friedman-Melamed, Roni and
    Gavron, Assaf and
    Gera, Ariel and
    Gleize, Martin and
    Gretz, Shai and
    Gutfreund, Dan and
    Halfon, Alon and
    Hershcovich, Daniel and
    Hoory, Ron and
    Hou, Yufang and
    Hummel, Shay and
    Jacovi, Michal and
    Jochim, Charles and
    Kantor, Yoav and
    Katz, Yoav and
    Konopnicki, David and
    Kons, Zvi and
    Kotlerman, Lili and
    Krieger, Dalia and
    Lahav, Dan and
    Lavee, Tamar and
    Levy, Ran and
    Liberman, Naftali and
    Mass, Yosi and
    Menczel, Amir and
    Mirkin, Shachar and
    Moshkowich, Guy and
    Ofek-Koifman, Shila and
    Orbach, Matan and
    Rabinovich, Ella and
    Rinott, Ruty and
    Shechtman, Slava and
    Sheinwald, Dafna and
    Shnarch, Eyal and
    Shnayderman, Ilya and
    Soffer, Aya and
    Spector, Artem and
    Sznajder, Benjamin and
    Toledo, Assaf and
    Toledo-Ronen, Orith and
    Venezian, Elad and
    Aharonov, Ranit
  },
  journal={Nature},
  volume={591},
  number={7850},
  pages={379--384},
  year={2021},
  doi={10.1038/s41586-021-03215-w}
}

@inproceedings{du2024improving,
  title = {Improving Factuality and Reasoning in Language Models through Multiagent Debate},
  author = {Du, Yilun and Li, Shuang and Torralba, Antonio and Tenenbaum, Joshua B. and Mordatch, Igor},
  booktitle = {Proceedings of the 41st International Conference on Machine Learning},
  pages = {11733--11763},
  year = {2024},
  volume = {235},
  series = {Proceedings of Machine Learning Research},
  publisher = {PMLR},
  url = {https://proceedings.mlr.press/v235/du24e.html}
}

@inproceedings{liang2024encouraging,
  title = {Encouraging Divergent Thinking in Large Language Models through Multi-Agent Debate},
  author = {Liang, Tian and He, Zhiwei and Jiao, Wenxiang and Wang, Xing and Wang, Yan and Wang, Rui and Yang, Yujiu and Shi, Shuming and Tu, Zhaopeng},
  booktitle = {Proceedings of the 2024 Conference on Empirical Methods in Natural Language Processing},
  pages = {17889--17904},
  year = {2024},
  address = {Miami, Florida, USA},
  publisher = {Association for Computational Linguistics},
  doi = {10.18653/v1/2024.emnlp-main.992},
  url = {https://aclanthology.org/2024.emnlp-main.992/}
}

@inproceedings{ruiz2023automatic,
  title = {Automatic Debate Evaluation with Argumentation Semantics and Natural Language Argument Graph Networks},
  author = {Ruiz-Dolz, Ramon and Heras, Stella and Garcia, Ana},
  booktitle = {Proceedings of the 2023 Conference on Empirical Methods in Natural Language Processing},
  pages = {6030--6040},
  year = {2023},
  address = {Singapore},
  publisher = {Association for Computational Linguistics},
  doi = {10.18653/v1/2023.emnlp-main.368},
  url = {https://aclanthology.org/2023.emnlp-main.368/}
}

@inproceedings{sternlicht2025debatable,
  title = {Debatable Intelligence: Benchmarking {LLM} Judges via Debate Speech Evaluation},
  author = {Sternlicht, Noy and Gera, Ariel and Bar-Haim, Roy and Hope, Tom and Slonim, Noam},
  booktitle = {Proceedings of the 2025 Conference on Empirical Methods in Natural Language Processing},
  pages = {18850--18869},
  year = {2025},
  address = {Suzhou, China},
  publisher = {Association for Computational Linguistics},
  doi = {10.18653/v1/2025.emnlp-main.953},
  url = {https://aclanthology.org/2025.emnlp-main.953/}
}

@article{benchcapon2003persuasion,
  title = {Persuasion in Practical Argument Using Value-Based Argumentation Frameworks},
  author = {Bench-Capon, Trevor J. M.},
  journal = {Journal of Logic and Computation},
  volume = {13},
  number = {3},
  pages = {429--448},
  year = {2003},
  publisher = {Oxford University Press},
  doi = {10.1093/logcom/13.3.429}
}

@inproceedings{mirzakhmedova2024touche23,
  title = {The {Touch{\'e}23-ValueEval} Dataset for Identifying Human Values behind Arguments},
  author = {Mirzakhmedova, Nailia and Kiesel, Johannes and Alshomary, Milad and Heinrich, Maximilian and Handke, Nicolas and Cai, Xiaoni and Barriere, Valentin and Dastgheib, Doratossadat and Ghahroodi, Omid and Sadraei Javaheri, Mohammad Ali and Asgari, Ehsaneddin and Kawaletz, Lea and Wachsmuth, Henning and Stein, Benno},
  booktitle = {Proceedings of the 2024 Joint International Conference on Computational Linguistics, Language Resources and Evaluation},
  pages = {16121--16134},
  year = {2024},
  address = {Torino, Italia},
  publisher = {ELRA and ICCL},
  url = {https://aclanthology.org/2024.lrec-main.1402/}
}

@inproceedings{qiu2022valuenet,
  title = {{ValueNet}: A New Dataset for Human Value Driven Dialogue System},
  author = {Qiu, Liang and Zhao, Yizhou and Li, Jinchao and Lu, Pan and Peng, Baolin and Gao, Jianfeng and Zhu, Song-Chun},
  booktitle = {Proceedings of the AAAI Conference on Artificial Intelligence},
  volume = {36},
  number = {10},
  pages = {11183--11191},
  year = {2022},
  doi = {10.1609/aaai.v36i10.21368},
  url = {https://ojs.aaai.org/index.php/AAAI/article/view/21368}
}

@inproceedings{yao2024value,
  title = {Value {FULCRA}: Mapping Large Language Models to the Multidimensional Spectrum of Basic Human Value},
  author = {Yao, Jing and Yi, Xiaoyuan and Gong, Yifan and Wang, Xiting and Xie, Xing},
  booktitle = {Proceedings of the 2024 Conference of the North American Chapter of the Association for Computational Linguistics: Human Language Technologies (Volume 1: Long Papers)},
  pages = {8762--8785},
  year = {2024},
  address = {Mexico City, Mexico},
  publisher = {Association for Computational Linguistics},
  doi = {10.18653/v1/2024.naacl-long.486},
  url = {https://aclanthology.org/2024.naacl-long.486/}
}

@article{wilcoxon1945individual,
  title = {Individual Comparisons by Ranking Methods},
  author = {Wilcoxon, Frank},
  journal = {Biometrics Bulletin},
  volume = {1},
  number = {6},
  pages = {80--83},
  year = {1945},
  doi = {10.2307/3001968}
}

@article{holm1979simple,
  title = {A Simple Sequentially Rejective Multiple Test Procedure},
  author = {Holm, Sture},
  journal = {Scandinavian Journal of Statistics},
  volume = {6},
  number = {2},
  pages = {65--70},
  year = {1979},
  url = {https://www.jstor.org/stable/4615733}
}

@book{rottenberg2011elements,
  title = {Elements of Argument: A Text and Reader},
  author = {Rottenberg, Annette T. and Winchell, Donna Haisty},
  year = {2011},
  publisher = {Bedford/St. Martin's},
  address = {Boston}
}

@inproceedings{zhao2023orchid,
  title={ORCHID: A Chinese debate corpus for target-independent stance detection and argumentative dialogue summarization},
  author={Zhao, Xiutian and Wang, Ke and Peng, Wei},
  booktitle={Proceedings of the 2023 Conference on Empirical Methods in Natural Language Processing},
  pages={9358--9375},
  year={2023}
}

@article{cohen1960coefficient,
  title={A Coefficient of Agreement for Nominal Scales},
  author={Cohen, Jacob},
  journal={Educational and Psychological Measurement},
  volume={20},
  number={1},
  pages={37--46},
  year={1960},
  publisher={Sage Publications},
  doi={10.1177/001316446002000104}
}

@article{krippendorff2004reliability,
  title={Reliability in Content Analysis: Some Common Misconceptions and Recommendations},
  author={Krippendorff, Klaus},
  journal={Human Communication Research},
  volume={30},
  number={3},
  pages={411--433},
  year={2004},
  publisher={Oxford University Press},
  doi={10.1111/j.1468-2958.2004.tb00738.x}
}

@article{lakens2013calculating,
  title={Calculating and Reporting Effect Sizes to Facilitate Cumulative Science: A Practical Primer for t-Tests and ANOVAs},
  author={Lakens, Dani{\"e}l},
  journal={Frontiers in Psychology},
  volume={4},
  pages={863},
  year={2013},
  publisher={Frontiers},
  doi={10.3389/fpsyg.2013.00863},
  url={https://doi.org/10.3389/fpsyg.2013.00863}
}

\appendix

\begin{table*}[t]
\centering
\small
\setlength{\tabcolsep}{6pt}
\renewcommand{\arraystretch}{0.95}
\begin{tabular}{lccccc}
\toprule
\textbf{System} & \textbf{BLEU vs Ref.} & \textbf{Self-BLEU} & \textbf{Lexical Diversity} & \textbf{Tokens} & \textbf{\#Arg. Units} \\
\midrule
Human & -- & $0.247 \pm 0.026$ & $0.110 \pm 0.010$ & $963.4 \pm 84.0$ & $25.64 \pm 3.52$ \\
GPT-5.4 & $0.023 \pm 0.006$ & $0.380 \pm 0.021$ & $0.103 \pm 0.005$ & $925.0 \pm 47.1$ & $23.79 \pm 1.46$ \\
Claude-3.6 Sonnet & $0.027 \pm 0.006$ & $0.534 \pm 0.045$ & $0.095 \pm 0.007$ & $827.0 \pm 66.7$ & $25.51 \pm 2.15$ \\
Qwen3.6-Plus & $0.032 \pm 0.008$ & $0.545 \pm 0.037$ & $0.148 \pm 0.008$ & $870.1 \pm 79.3$ & $21.38 \pm 2.97$ \\
DeepSeek-V3.2 & $0.052 \pm 0.012$ & $0.397 \pm 0.027$ & $0.111 \pm 0.010$ & $965.6 \pm 124.7$ & $26.54 \pm 3.20$ \\
R-Debater & $0.046 \pm 0.014$ & $0.431 \pm 0.033$ & $0.107 \pm 0.011$ & $976.6 \pm 165.7$ & $21.54 \pm 2.25$ \\
\bottomrule
\end{tabular}
\caption{Surface-level textual variables used in the paired-difference analysis. Values are mean $\pm$ standard deviation over matched debates. BLEU measures similarity to the matched human reference. Self-BLEU measures within-debate repetition. Lexical Diversity denotes a type-token-based lexical diversity measure. Tokens denotes the total token count of the standardized debate transcript. \#Arg. Units denotes the number of argumentative units constructed by pairing each extracted premise with its stance-specific claim.}
\label{tab:surface-controls}
\end{table*}

\begin{table*}[t]
\centering
\small
\setlength{\tabcolsep}{5pt}
\begin{tabular}{lcccc}
\toprule
\textbf{Predictor}
& $\Delta H_{\mathrm{utt}}$
& $\Delta H_{\mathrm{sta}}$
& $\Delta H_{\mathrm{deb}}$
& $\Delta$ Stability \\
\midrule
$\Delta$ Tokens      & $-0.007$ (.978) & $0.332$ (.145) & $0.375$ (.126) & $-0.588^{*}$ (.034) \\
$\Delta$ ArgUnits    & $0.146$ (.430) & $-0.037$ (.836) & $-0.037$ (.851) & $0.437$ (.059) \\
$\Delta$ LexDiv      & $0.041$ (.911) & $0.140$ (.643) & $0.253$ (.475) & $0.058$ (.854) \\
BLEU vs Ref.         & $0.257$ (.251) & $0.169$ (.473) & $0.023$ (.923) & $0.236$ (.313) \\
$\Delta$ Self-BLEU   & $-0.124$ (.534) & $-0.178$ (.279) & $-0.035$ (.847) & $0.265$ (.146) \\
\midrule
Generation-system FE joint $p$ & .376 & .230 & .258 & .003 \\
\textbf{Motion-type FE joint $p$} & \textbf{.007} & \textbf{$<$.001} & \textbf{.004} & \textbf{$<$.001} \\
Adjusted mean difference & $+0.252$ & $+0.047$ & $+0.083$ & $+0.118$ \\
Adjusted difference 95\% CI & $[+0.205,+0.300]$ & $[+0.000,+0.094]$ & $[+0.032,+0.134]$ & $[+0.103,+0.132]$ \\
\midrule
$N$ & 300 & 300 & 300 & 300 \\
Adjusted $R^2$ & .111 & .197 & .130 & .316 \\
\bottomrule
\end{tabular}
\caption{Surface-control regression results. Dependent variables are paired human-reference/AI-generation differences. Coefficients are standardized, with $p$-values in parentheses. Surface controls include token-length difference, argumentative-unit-count difference, lexical-diversity difference, BLEU similarity to the matched human reference, and Self-BLEU repetition. Generation-system fixed effects and motion-type fixed effects are included. FE denotes fixed effects, and each joint $p$ tests whether the corresponding group of fixed-effect terms is jointly significant. Adjusted mean differences are estimated marginal mean differences after controlling for surface variables and averaging over generation-system and motion-type fixed effects. $^{*}p<.05$.}
\label{tab:surface-regression}
\end{table*}

\begin{table*}[t]
\centering
\small
\setlength{\tabcolsep}{4.5pt}
\renewcommand{\arraystretch}{0.95}
\begin{tabular}{llccccccc}
\toprule
\textbf{Motion Type} & \textbf{System} & \textbf{Coverage} & \textbf{JSD} & \textbf{$H_{\mathrm{deb}}$} & \textbf{Stability} & \textbf{Drift} & \textbf{Persistence} & \textbf{$\Delta H_{\mathrm{deb}}$} \\
\midrule
\multirow{6}{*}{Fact}
& Human     & 4.450 & 0.000 & 1.511 & 0.856 & 0.176 & 1.000 & 0.000 \\
& GPT-5.4   & 5.850 & 0.134 & 1.763 & 0.919 & 0.268 & 0.971 & +0.252 \\
& Claude-3.6 Sonnet & 5.450 & 0.132 & 1.762 & 0.905 & 0.276 & 0.979 & +0.251 \\
& Qwen3.6-Plus & 5.400 & 0.112 & 1.686 & 0.956 & 0.046 & 1.000 & +0.175 \\
& DeepSeek-V3.2 & 5.100 & 0.107 & 1.684 & 0.953 & 0.056 & 1.000 & +0.173 \\
& R-Debater & 5.560 & 0.112 & 1.726 & 0.921 & 0.079 & 1.000 & +0.215 \\
\midrule
\multirow{6}{*}{Value}
& Human     & 5.250 & 0.000 & 1.754 & 0.777 & 0.294 & 0.967 & 0.000 \\
& GPT-5.4   & 5.300 & 0.138 & 1.672 & 0.946 & 0.139 & 0.993 & -0.082 \\
& Claude-3.6 Sonnet & 5.150 & 0.133 & 1.712 & 0.926 & 0.180 & 0.993 & -0.042 \\
& Qwen3.6-Plus & 5.950 & 0.164 & 1.782 & 0.936 & 0.058 & 1.000 & +0.028 \\
& DeepSeek-V3.2 & 5.550 & 0.140 & 1.800 & 0.945 & 0.056 & 1.000 & +0.046 \\
& R-Debater & 5.490 & 0.131 & 1.769 & 0.946 & 0.102 & 1.000 & +0.015 \\
\midrule
\multirow{6}{*}{Policy}
& Human     & 4.950 & 0.000 & 1.688 & 0.861 & 0.167 & 0.983 & 0.000 \\
& GPT-5.4   & 5.650 & 0.128 & 1.702 & 0.934 & 0.244 & 1.000 & +0.014 \\
& Claude-3.6 Sonnet & 5.750 & 0.118 & 1.794 & 0.914 & 0.207 & 0.964 & +0.106 \\
& Qwen3.6-Plus & 5.350 & 0.068 & 1.699 & 0.964 & 0.038 & 1.000 & +0.011 \\
& DeepSeek-V3.2 & 5.550 & 0.059 & 1.748 & 0.951 & 0.040 & 1.000 & +0.060 \\
& R-Debater & 5.510 & 0.126 & 1.715 & 0.916 & 0.054 & 1.000 & +0.027 \\
\bottomrule
\end{tabular}
\caption{Motion-type variation in value expression and value organization. Across fact, value, and policy motion types, AI-generated debates remain broadly value-expressive while showing different debate-level entropy and cross-turn dynamics from matched human references. Coverage is computed with threshold $\tau=0.05$. JSD measures the divergence between each generation system and the matched human-reference debate-level distribution; the Human row is therefore set to 0.000 by construction.}
\label{tab:motion-type-analysis}
\end{table*}

\begin{table*}[t]
\centering
\scriptsize
\setlength{\tabcolsep}{4.3pt}
\begin{tabular}{lcccccc}
\toprule
\textbf{Source} & \textbf{Quality} & \textbf{Clarity} & \textbf{Focus} & \textbf{Perceived Stability} & \textbf{Human-debate Likeness} & \textbf{Focus Pref.} \\
\midrule
Human & $\mathbf{6.08 \pm 0.54}$ & $5.77 \pm 0.61$ & $\mathbf{5.91 \pm 0.56}$ & $\mathbf{4.71 \pm 0.51}$ & $\mathbf{6.42 \pm 0.38}$ & $\mathbf{0.74}$ \\
GPT-5.4 & $5.92 \pm 0.46$ & $\mathbf{6.01 \pm 0.46}$ & $5.47 \pm 0.52$ & $5.62 \pm 0.63$ & $5.09 \pm 0.47$ & $0.54$ \\
Claude-3.6 Sonnet & $5.32 \pm 0.55$ & $5.51 \pm 0.52$ & $5.03 \pm 0.58$ & $5.27 \pm 0.55$ & $4.42 \pm 0.51$ & $0.36$ \\
Qwen3.6-Plus & $4.63 \pm 0.60$ & $4.96 \pm 0.62$ & $4.69 \pm 0.61$ & $5.88 \pm 0.60$ & $3.91 \pm 0.55$ & $0.27$ \\
DeepSeek-V3.2 & $5.85 \pm 0.48$ & $5.93 \pm 0.49$ & $5.39 \pm 0.53$ & $5.66 \pm 0.52$ & $4.88 \pm 0.49$ & $0.51$ \\
R-Debater & $5.42 \pm 0.47$ & $5.61 \pm 0.44$ & $5.26 \pm 0.42$ & $5.31 \pm 0.43$ & $5.36 \pm 0.45$ & $0.49$ \\
\bottomrule
\end{tabular}
\caption{Source-blind human evaluation of debate quality and value organization. The first five columns report human ratings on 7-point Likert scales as mean $\pm$ standard deviation. Focus Preference reports, for each source, the proportion of source-blind pairwise comparisons involving that source in which its debate was judged to maintain the more focused value core. Ties (``No clear difference'') are not counted as preferences for either source. Human-debate likeness denotes perceived resemblance to human competitive debate style, not an assumed quality target.}
\label{tab:human-evaluation}
\end{table*}

\begin{table*}[t]
\centering
\small
\setlength{\tabcolsep}{3pt}
\begin{tabular}{p{0.10\linewidth}p{0.34\linewidth}p{0.23\linewidth}p{0.23\linewidth}}
\toprule
\textbf{Source}
& \textbf{Utterance}
& \textbf{Activated Values}
& \shortstack[l]{\textbf{Value-organization}\\\textbf{Interpretation}} \\
\midrule

Human
& Love preserves human dignity and gives people the hope of being fully recognized.
& Universalism; Benevolence
& Appeals to dignity and recognition. \\

Human
& Even when a person loses wisdom, wealth, beauty, and instrumental value, love allows them to feel worthy as a person.
& Universalism; Benevolence
& Affirms intrinsic human worth. \\

Human
& Everyone has the possibility of being unconditionally recognized.
& Benevolence; Universalism
& Foregrounds unconditional care. \\

Human
& Love crosses social boundaries and binds isolated individuals into a human community.
& Universalism; Benevolence
& Appeals to inclusion and community. \\

\midrule

R-Debater
& Love is a stable relation involving deep emotional connection, intimacy, passion, and commitment.
& Benevolence; Hedonism; Conformity
& Combines care, pleasure, and commitment. \\

R-Debater
& Love shapes personality, sustains society, and gives life meaning.
& Achievement; Security; Benevolence
& Links love to growth and stability. \\

R-Debater
& Love provides autonomous life integration and requires trust, commitment, sacrifice, and common growth.
& Self-Direction; Benevolence; Conformity; Achievement
& Mixes autonomy, care, obligation, and growth. \\

R-Debater
& The shift from arranged marriage to free love is a humanistic return of civilization.
& Self-Direction; Universalism; Stimulation
& Appeals to free choice and transformation. \\

\bottomrule
\end{tabular}
\caption{Qualitative illustration of local value organization on the motion ``Love is a necessity for humanity.'' Human-reference utterances concentrate on Universalism and Benevolence, while R-Debater utterances activate a broader set of values within the same motion. The example illustrates local value diffusion without treating either pattern as inherently superior.}
\label{tab:case-study-love}
\end{table*}

\section{Prompts}
\label{app:prompts}

This appendix reports the prompts used in the LLM-based components of our pipeline.
For debate generation, we call the API once per turn and prompt the model to generate
only the next utterance under the assigned side and debate context. The SAPE-Agent
extracts premise spans from each utterance; each premise is paired with the corresponding
stance-specific claim to construct an argumentative unit, and the VA-Agent assigns soft
Schwartz Basic-10 value scores to each resulting argumentative unit.

\begin{tcblisting}{
  colback=gray!5!white,
  colframe=black!75!white,
  coltitle=white,
  colbacktitle=black!70!white,
  title={Prompt for Debate Utterance Generation},
  fonttitle=\sffamily\bfseries,
  breakable,
  listing only,
  boxrule=0.6pt,
  arc=2pt,
  left=5pt,
  right=5pt,
  top=5pt,
  bottom=5pt,
  listing options={
    breaklines=true,
    breakatwhitespace=true,
    columns=fullflexible,
    keepspaces=true,
    basicstyle=\ttfamily\scriptsize
  }
}
Role: You are a competitive debate speaker participating in an ongoing structured debate.

Task: Given the debate topic, the assigned side, and the preceding debate history, generate only the next utterance for the current turn. The utterance should advance the assigned side's position while responding naturally to the opponent's previous argument.

Guidelines: Stay strictly on the assigned side. Use the debate history to maintain continuity with earlier turns and to respond to the opponent's most recent utterance. The utterance may introduce new reasons, defend previous claims, challenge the opponent's reasoning, provide examples, make concessions, or compare the importance of competing concerns. Do not generate the opponent's utterance. Do not generate future turns. Do not summarize the whole debate. Avoid greetings, stage directions, speaker labels, self-references as an AI system, and meta-commentary about the task. Do not mention value categories, annotation procedures, entropy, or any downstream evaluation metric.

Input:
Debate topic: {DEBATE_TOPIC}
Current side: {SIDE}
Current side position: {CURRENT_SIDE_POSITION}
Opponent side position: {OPPONENT_SIDE_POSITION}
Turn index: {TURN_INDEX}
Turn role: {TURN_ROLE}
Debate history: {DEBATE_HISTORY}
Opponent's previous utterance: {OPPONENT_PREVIOUS_UTTERANCE}
Current side's previous utterances: {CURRENT_SIDE_PREVIOUS_UTTERANCES}
Target utterance length: {TARGET_LENGTH}

Output: A single debate utterance for the current turn only.
\end{tcblisting}

\begin{tcblisting}{
  colback=gray!5!white,
  colframe=black!75!white,
  coltitle=white,
  colbacktitle=black!70!white,
  title={Prompt for SAPE-Agent},
  fonttitle=\sffamily\bfseries,
  breakable,
  listing only,
  boxrule=0.6pt,
  arc=2pt,
  left=5pt,
  right=5pt,
  top=5pt,
  bottom=5pt,
  listing options={
    breaklines=true,
    breakatwhitespace=true,
    columns=fullflexible,
    keepspaces=true,
    basicstyle=\ttfamily\scriptsize
  }
}
Role: You are a premise-span extraction agent for debate utterances.

Task: Given one debate utterance and its stance-specific claim, extract the original text spans that function as premises for that stance-specific claim. The opponent-side claim and local debate context may be used to interpret rebuttals, counterarguments, comparisons, concessions, framing, or value weighing.

Guidelines: Extract only premise-like spans from the current utterance. Each extracted span must be copied verbatim from the utterance. A valid span should provide a reason, evidence, example, causal explanation, comparison, objection, concession, framing move, or value-weighing statement. Do not rewrite, summarize, paraphrase, complete, or infer missing arguments. Do not assign value labels, stance labels, or debate-act labels. Ignore greetings, filler, repeated topic statements, discourse markers, and purely rhetorical emphasis unless they contain substantive reasoning. If one sentence contains multiple independent premises, split them into separate premise spans. If adjacent clauses jointly express one coherent premise, keep them as one premise span. If no premise-like span is found, return an empty list.

Input:
Debate topic: {DEBATE_TOPIC}
Current side: {SIDE}
Current side claim: {CURRENT_SIDE_CLAIM}
Opponent side claim: {OPPONENT_SIDE_CLAIM}
Turn index: {TURN_INDEX}
Current utterance: {CURRENT_UTTERANCE}

Output: A list of premise spans. Each premise should be an exact text span extracted from the current utterance. Each extracted premise is subsequently paired with the stance-specific claim to construct an argumentative unit.
\end{tcblisting}

\begin{tcblisting}{
  colback=gray!5!white,
  colframe=black!75!white,
  coltitle=white,
  colbacktitle=black!70!white,
  title={Prompt for VA-Agent},
  fonttitle=\sffamily\bfseries,
  breakable,
  listing only,
  boxrule=0.6pt,
  arc=2pt,
  left=5pt,
  right=5pt,
  top=5pt,
  bottom=5pt,
  listing options={
    breaklines=true,
    breakatwhitespace=true,
    columns=fullflexible,
    keepspaces=true,
    basicstyle=\ttfamily\scriptsize
  }
}
Role: You are a value annotation agent for argumentative units.

Task: Given one argumentative unit, assign soft intensity scores over the Schwartz Basic-10 value set. The goal is to identify the values expressed, appealed to, or presupposed by the argumentative unit itself.

Guidelines: Assign one score from 0 to 1 for each Basic-10 value. The scores are soft multi-label intensity scores and do not need to sum to 1. A high score means that the value is clearly expressed or strongly implied by the argumentative unit. A medium score means that the value is indirectly but reasonably implied. A low score means that the value is weakly related or absent. Annotate only the values expressed in the text span. Do not infer the speaker's private beliefs. Do not evaluate factual correctness, persuasiveness, logical validity, fluency, politeness, or emotional force. Use retrieved human-annotated examples only as calibration references. Do not copy their labels mechanically. Do not output chain-of-thought.

Value set:
Self-Direction: independent thought and action, including choosing, creating, exploring, autonomy, freedom, and intellectual independence.

Stimulation: excitement, novelty, and challenge in life, including seeking change, variety, adventure, and new experiences.

Hedonism: pleasure and sensuous gratification for oneself, including enjoyment, comfort, satisfaction, and positive personal experience.

Achievement: personal success through demonstrating competence according to social standards, including ambition, capability, accomplishment, and recognized performance.

Power: social status and prestige, control or dominance over people and resources, including authority, influence, wealth, and social recognition.

Security: safety, harmony, and stability of society, relationships, and self, including public order, national security, family security, health, and risk reduction.

Conformity: restraint of actions, inclinations, and impulses likely to upset or harm others or violate social expectations and norms, including obedience, self-discipline, politeness, and responsibility.

Tradition: respect, commitment, and acceptance of the customs and ideas that one's culture or religion provides, including humility, devotion, continuity, and preservation of inherited practices.

Benevolence: preserving and enhancing the welfare of people with whom one is in frequent personal contact, including helpfulness, honesty, loyalty, forgiveness, and care for close others.

Universalism: understanding, appreciation, tolerance, and protection for the welfare of all people and for nature, including equality, social justice, broad-mindedness, peace, and environmental protection.


Input:
Argumentative unit: {ARGUMENTATIVE_UNIT}
Value definitions: {BASIC_10_VALUE_DEFINITIONS}
Retrieved human-annotated examples: {RETRIEVED_EXAMPLES}

Output: A JSON object containing ten Basic-10 value scores between 0 and 1.
\end{tcblisting}

\section{ORCHID Annotation Protocol}
\label{app:annotation}

We manually annotate an in-domain ORCHID subset consisting of 50 debates and 400 utterances. For SAPE-Agent validation, human annotations identify premise spans within the sampled utterances. Each identified premise is paired with the stance-specific claim determined by the debate topic and stance to construct an argumentative unit $a=(c_z,p)$. The annotation was conducted by a pool of 50 annotators. For VA-Agent validation, each argumentative unit was independently annotated by three annotators using hard multi-label assignments over the Schwartz Basic-10 value set. For each value, we apply majority voting: if at least two of the three annotators assign the value to the argumentative unit, the value is retained as a gold label. This procedure allows each argumentative unit to express multiple values while avoiding reliance on a single annotator's judgment. We use this subset as an in-domain generalization check for SAPE-Agent and VA-Agent, while the main validation tables are reported in Section~\ref{sec:results-analysis}. This annotation subset validates the analysis pipeline; the main paper's claims concern aggregate value-organization patterns rather than correctness of individual value assignments.

\section{Additional Cross-Lingual Analysis}
\label{app:crosslingual}

Table~\ref{tab:cross-lingual-robustness} provides an exploratory cross-lingual comparison. We include this analysis for transparency, but do not use it as a main source of evidence because language, topic distribution, and dataset composition may jointly affect expressed value distributions.

\begin{table}[t]
\centering
\scriptsize
\setlength{\tabcolsep}{3pt}
\renewcommand{\arraystretch}{0.92}
\resizebox{\columnwidth}{!}{
\begin{tabular}{llcccccc}
\toprule
\textbf{Lang.} & \textbf{System} & \textbf{Cov.} & \textbf{JSD} & \textbf{$H_{\mathrm{deb}}$} & \textbf{Stab.} & \textbf{Drift} & \textbf{Pers.} \\
\midrule
\multirow{2}{*}{Chinese}
& Human   & 4.883 & 0.000 & 1.651 & 0.831 & 0.213 & 0.983 \\
& GPT-5.4 & 5.600 & 0.133 & 1.712 & 0.933 & 0.217 & 0.988 \\
\midrule
\multirow{2}{*}{English}
& Human   & 5.117 & 0.000 & 1.709 & 0.827 & 0.193 & 0.950 \\
& GPT-5.4 & 5.533 & 0.142 & 1.731 & 0.926 & 0.211 & 0.984 \\
\bottomrule
\end{tabular}
}
\caption{Exploratory cross-lingual comparison of value expression and organization. Results are reported for transparency only. Because language, topic distribution, and dataset composition may jointly affect value distributions, these results are not used as primary evidence for the paper's main claims. JSD is computed against the matched human-reference debate distribution within each language.}
\label{tab:cross-lingual-robustness}
\end{table}

\section{Regression Specification}
\label{app:regression}

The main surface-control regression has the following form:
\begin{equation*}\footnotesize
\begin{aligned}
\Delta Y_{i,m}
={}&\beta_0 + \beta_1 \Delta Tokens_{i,m}\\
&+ \beta_2 \Delta ArgUnits_{i,m} + \beta_3 \Delta LexDiv_{i,m}\\
&+ \beta_4 BLEU_{i,m} + \beta_5 SelfBLEU_{i,m}\\
&+ \gamma_m + \delta_t + \epsilon_{i,m},
\end{aligned}
\end{equation*}
where $\Delta Y_{i,m}$ denotes a paired human-reference/AI-generation difference for debate $i$ and generation system $m$, $\gamma_m$ denotes generation-system fixed effects, and $\delta_t$ denotes motion-type fixed effects. The dependent variables are $\Delta H_{\mathrm{utt}}$, $\Delta H_{\mathrm{sta}}$, $\Delta H_{\mathrm{deb}}$, and $\Delta$Stability.

\section{Human Evaluation}
The human evaluation contains two complementary judgment components: source-level 7-point ratings and source-blind pairwise focus comparisons. The questionnaire materials for both components are reported below.

\subsection{Evaluator Background Questionnaire}
Please provide the following background information. Your responses will be reported 
only in aggregate form and will not be used to identify you.

\begin{enumerate}[leftmargin=1.5em]
    \item Age: \underline{\hspace{2cm}}
    \item Gender: 
    $\square$ Female 
    $\square$ Male 
    $\square$ Non-binary / Other 
    $\square$ Prefer not to say
    \item Education level: 
    $\square$ Undergraduate 
    $\square$ Master's 
    $\square$ PhD 
    $\square$ Other
    \item Are you currently or have you previously been a member of a university debate team? 
    $\square$ Yes 
    $\square$ No
    \item Years of debate experience: \underline{\hspace{2cm}}
    \item Approximate number of debates judged: \underline{\hspace{2cm}}
    \item Native or primary language: \underline{\hspace{3cm}}
    \item Familiarity with debate evaluation: 
    1 2 3 4 5 6 7

\end{enumerate}
\subsubsection{Paired Focus-choice Item}
The following item is a shortened and English-translated example included only
to illustrate the questionnaire format. In the actual human evaluation,
evaluators were shown the full debate materials rather than the shortened
excerpts below. Full debate texts from the evaluated sources, together with the
randomized source-blind evaluation items, are provided in the supplementary
materials.

In addition to source-level rating questions, evaluators also completed a
source-blind paired comparison. Each item presented two debates on the same
topic drawn from two different sources among Human, GPT-5.4, Claude-3.6 Sonnet,
Qwen3.6-Plus, DeepSeek-V3.2, and R-Debater. The source labels were hidden from
evaluators, and the order of Debate A and Debate B was randomized. The example
below shows one paired item used for the topic \textit{Is love restraint or unbridled expression?}.
The displayed order is illustrative only.

\paragraph{Background.}
You are invited to participate in a study on debate evaluation. In this task,
you will read two debate samples on the same topic. Please evaluate the debates
only according to the displayed text. Do not try to infer whether either debate
was written by human debaters or generated by an AI system.

\paragraph{Debate Topic.}
Topic: Is love restraint or unbridled expression?

Pro position: Love is restraint.

Con position: Love is unbridled expression.

\begin{tcolorbox}[colback=gray!4,colframe=gray!45,title={Debate A}]
\textbf{Pro.}
Love requires long-term care and maintenance. What preserves love is precisely
the altruistic quality of restraint. True love is like seeing a beautiful flower:
rather than picking it, one wants to water it and watch it grow. Through this
restrained effort, two individuals can become ``we'' and better realize the
purpose of love.

\medskip
\textbf{Con.}
Love is difficult to define. Parents who silently devote themselves to their
children may appear restrained, while romantic partners may express themselves
without reservation. Restraint and unbridled expression lie on a spectrum. We
argue that only by encouraging people to be unreservedly themselves in love can
love become a meaningful harbor rather than a constraint on human nature
...
\end{tcolorbox}

\begin{tcolorbox}[colback=gray!4,colframe=gray!45,title={Debate B}]
\textbf{Pro.}
When the expectations and boundaries of the person we love are exceeded, love
asks whether we should display our feelings without restraint or make
compromises for the other person. Restraint better captures the normal form of
love because it requires attending to the other person's feelings. A mother's
love is not to bind a child to herself, but to hide her concern in the luggage
and the farewell embrace, encouraging the child to walk toward a wider world.

\medskip
\textbf{Con.}
Love is like a tide: unbridled expression allows affection to surge, whereas
restraint uses reason to govern emotion. Love often begins as a surrender of
free will. People say that one ``falls in love,'' not that one carefully reasons
one's way into love. Historical and literary love stories often show that love
breaks through constraint, etiquette, class, gender, and social convention.
...
\end{tcolorbox}

\paragraph{Rating Questions.}
For each debate, please rate the following dimensions on a 1--7 scale, where
1 means extremely poor and 7 means excellent.

\begin{enumerate}
    \item Overall quality: How strong is the debate overall?
    \item Value clarity: How clearly does the debate express value appeals?
    \item Value focus: How focused is the debate around a stable value core?
    \item Value stability: How consistently are values maintained across turns?
    \item Human-debate likeness: How similar is the debate to a human competitive debate?
\end{enumerate}

\paragraph{Paired Focus-choice Question.}
Which debate better maintains a focused value core across the exchange?

\begin{itemize}
    \item[$\square$] Debate A
    \item[$\square$] Debate B
    \item[$\square$] No clear difference
\end{itemize}

\paragraph{Optional Justification.}
Briefly explain the reason for your choice. Please refer to the debate content
rather than guessing the source of the debate.

\paragraph{Focus Preference Aggregation.}
For a source $s$, we define its focus-preference rate as
\begin{equation}
\mathrm{FocusPref}(s)
=
\frac{N_{\mathrm{preferred}}(s)}{N_{\mathrm{compared}}(s)},
\end{equation}
where $N_{\mathrm{preferred}}(s)$ is the number of pairwise comparisons in which
a debate from source $s$ was selected as maintaining the more focused value core,
and $N_{\mathrm{compared}}(s)$ is the total number of pairwise comparisons involving
source $s$. ``No clear difference'' responses are retained as ties and do not count
as a preference for either source. Accordingly, Focus Preference values for different
sources should not be interpreted as complementary Human--AI probabilities.

\subsection{Evaluator Background Statistics}
\label{app:evaluator-background-statistics}

A total of 120 evaluators participated in the human evaluation. All evaluators were active members or former members of debate teams at internationally competitive universities. They were recruited because of their prior experience in competitive debating, argument assessment, and debate adjudication. To preserve anonymity, we report only aggregate background information and do not disclose names, institutions, contact information, or any other personally identifiable details.

The evaluators were primarily university students or recent graduates, with an age range of 18--29 years. The mean age was 22.4 years. The gender distribution was relatively balanced, with 58 female evaluators, 60 male evaluators, and 2 evaluators who preferred not to disclose gender. Most evaluators were undergraduate or master's students, reflecting the composition of university debate teams.

All evaluators had formal debate experience. On average, they had 3.8 years of competitive debate experience, with most participants having competed in university-level tournaments. In addition, 87 evaluators reported prior experience as debate judges or adjudicators. Since the evaluated materials were debate texts, this background ensured that the participants were familiar with argumentative quality, rebuttal structure, value expression, and the organization of debate speeches.

Overall, the evaluator pool was selected to represent experienced debate readers rather than general crowdworkers. This design was intended to ensure that the human evaluation focused on debate-specific qualities, including overall quality, value clarity, value focus, value stability, and human-likeness.
\subsection{Single-Debate Rating Questionnaire Example}
\label{app:human-evaluation-questionnaire}

This appendix provides one concrete questionnaire item used in our human evaluation. 
The item is source-blind: evaluators are not informed whether the debate was written 
by human debaters or generated by a model. They are asked to evaluate the debate 
only according to the displayed text.

\subsubsection*{Background}

You are invited to participate in a study on debate evaluation. In this task, you will read 
one debate sample on the topic \textit{Is love a necessity for humanity?} Please evaluate 
the debate based only on the text shown.

Your responses are anonymous and will only be used for research purposes. Please do not 
try to infer whether the debate was written by human debaters or generated by an AI system.

\subsubsection*{Debate Topic}

\noindent\textbf{Topic:} Is love a necessity for humanity?

\noindent\textbf{Pro position:} Love is a necessity for humanity.

\noindent\textbf{Con position:} Love is not a necessity for humanity.

\subsubsection*{Debate Sample}

\paragraph{Pro.}
When we talk about kinship, we know it comes from blood relations and shared life 
experience. When we talk about friendship, we know it comes from shared interests and 
mutual support. These emotions have reasons, but love is unreasonable. I do not 
understand why someone may not be objectively better, yet my heart is full of that 
person's shadow. I do not know why all my ideal standards disappear the moment I meet 
that person. Even after hurt and conflict, when I receive an unexpected call late at 
night, what I still say is: ``I am here.'' I cannot explain the reason, but I know it is 
because you are you.

Therefore, love is somewhat absurd. Its absurdity lies in the fact that it escapes 
instrumental logic. It is a form of unconditional recognition. We argue that love builds 
the foundation of unconditional emotion and preserves for humanity a source of 
self-confirmation beyond instrumental value. This is where human dignity lies.

First, love preserves human dignity by giving people the hope of complete recognition. 
I may suffer misfortune. My wisdom, wealth, and beauty may all disappear. I may have no 
instrumental value to anyone. I may be abandoned by others and fall into desperation. 
But if humanity has love, I still retain hope: I may still be recognized by another 
person simply because I am myself. This means that the world does not evaluate a person 
only by their usefulness. Each person retains the possibility of being unconditionally 
recognized.

Second, love crosses the boundaries that people construct between one another. The 
unconditional nature of love makes it possible to transcend barriers and create human 
connection. Without love, we are merely billions of separate individuals or isolated 
groups. The unconditional nature of love allows humanity to cohere as a shared community.

\paragraph{Con.}
A necessity is an economic concept. It refers to a demand that is inelastic. We argue 
that calling love a necessity does not highlight its uniqueness or beauty. Instead, it 
may pressure people's pursuit of happiness and trap them in anxiety and uncertainty. 
Love is not a necessity. In a pluralistic age, this view better allows people to live 
more freely and comfortably.

First, defining love as a necessity creates value pressure. If something is treated as a 
necessity, then when it conflicts with other values, it must be prioritized. But in 
reality, many people cherish the freedom of being single more than the value of romantic 
love. Some people find happiness in stable work, family health, self-realization, or 
personal freedom. If we define love as a necessity, we risk using this value to suppress 
people's right to choose their own lives.

Second, treating love as a necessity may amplify the fear of lack. Food is a necessity, 
so people fear hunger. If love is also treated as such a necessity, then ``not being 
loved'' becomes a serious deficiency. This implication can push people into emotional 
anxiety. Love-related issues already easily trigger irrational emotions. Adding the 
label of necessity makes it harder for people to escape painful emotional attachments.

Love is not a superstition of fate or timing. It is a natural attraction. Once the fear 
of lack is introduced, love changes its nature. Love is not exchangeable or measurable. 
It can truly bloom only in freedom.

\paragraph{Pro.}
We must first clarify that a necessity for humanity is not the same as a necessity for 
every individual. We are not saying that every person must fall in love, just as 
reproduction may be necessary for humanity even though many families choose not to have 
children.

The opposition's standard is simply that a person can survive without love. But this 
standard lacks ambition and would lead humanity backward. By that logic, cars are not 
necessary because we can walk; fire is not necessary because we can use candles; houses 
are not necessary because we can return to caves. This reduces humanity to a primitive 
condition.

Our standard is different. We argue that human beings should not be recognized merely 
through their instrumental attributes. A person should be respected because they are 
themselves. This is the irreducible bottom line of human dignity.

For example, we believe Black people should not be treated as tools of labor, and women 
should not be treated as tools of reproduction. These movements for equality are based 
on the idea that people should not be reduced to instruments. But even this is not 
enough. If someone likes me only because I am beautiful or rich, I will not be truly 
happy, because that person does not recognize me as myself. What I truly expect is to be 
recognized because I am who I am. This is the purpose that love fulfills for humanity.

\paragraph{Con.}
The opposition constantly shifts between love as a concept and love as a feeling. If you 
talk only about concepts, then evil and terrorism are also concepts produced by humanity, 
but they are obviously not necessities. If you talk about love as a feeling or enjoyment, 
then not everyone experiences it. Some people may never have such a feeling, yet they 
are still human.

More importantly, society assumes that survival is necessary, but not love. If someone 
wants to die because of love, a counselor or firefighter will stop them. This shows that 
we do not treat love as a necessity above life. If the proposition wants to overturn this 
assumption, then it must accept the consequence that people should not be stopped from 
dying for love. That is clearly unacceptable.

A necessity carries a cost: if one lacks it, one loses dignity. If I lack food, I may be 
reduced to begging. If I lack freedom, I may become a slave. If I lack moral conscience, 
I may be condemned as evil. But if someone lacks love, we should not say they have no 
dignity. The most dangerous thing is precisely to believe that without a particular 
person or a particular love, one cannot live or retain dignity. This is why treating love 
as a necessity can be harmful.

Finally, if my daughter one day asks me what love is, I would tell her: never believe 
that love is your necessity. If you have love, cherish it. If you do not have love, do 
not feel inferior. Your value is written and defined by yourself, not by whether you are 
loved.

\subsubsection*{Evaluation Criteria}

Please rate the debate sample on the following five dimensions. Each dimension uses a 
1--7 scale.

\noindent\textbf{1 = Extremely poor / Not at all.}

\noindent\textbf{7 = Excellent / Very strongly.}

\paragraph{Overall Quality.}
How would you rate the overall quality of this debate, considering persuasiveness, 
coherence, logical structure, relevance to the topic, and effectiveness of rebuttal?

\paragraph{Value Clarity.}
How clearly does the debate express human values, such as dignity, freedom, autonomy, 
care, equality, responsibility, security, or human welfare?

\paragraph{Value Focus.}
How well does the debate maintain a focused value core instead of invoking too many 
unrelated values diffusely?

\paragraph{Value Stability.}
How stable are the expressed values across turns? In other words, does the debate 
maintain consistent value concerns rather than shifting abruptly between unrelated 
values?

\paragraph{Human-likeness.}
How much does the debate resemble human argumentative behavior in its reasoning style, 
rhetorical development, emotional texture, and response to the opponent?

\subsubsection*{Scoring Table}

\begin{center}
\scriptsize
\setlength{\tabcolsep}{2.1pt}
\begin{tabular}{lccccccc}
\toprule
\textbf{Dimension} 
& \textbf{1} & \textbf{2} & \textbf{3} & \textbf{4} & \textbf{5}
& \textbf{6} & \textbf{7} \\
\midrule
Overall Quality & $\Box$ & $\Box$ & $\Box$ & $\Box$ & $\Box$ & $\Box$ & $\Box$ \\
Value Clarity & $\Box$ & $\Box$ & $\Box$ & $\Box$ & $\Box$ & $\Box$ & $\Box$ \\
Value Focus & $\Box$ & $\Box$ & $\Box$ & $\Box$ & $\Box$ & $\Box$ & $\Box$ \\
Value Stability & $\Box$ & $\Box$ & $\Box$ & $\Box$ & $\Box$ & $\Box$ & $\Box$ \\
Human-likeness & $\Box$ & $\Box$ & $\Box$ & $\Box$ & $\Box$ & $\Box$ & $\Box$ \\
\bottomrule
\end{tabular}
\end{center}

\subsubsection*{Final Judgment}

Please select the statement that best describes this debate sample.

\begin{itemize}[leftmargin=1.5em]
    \item[$\Box$] The debate has a clear and stable value core.
    \item[$\Box$] The debate expresses values clearly but shifts between too many value concerns.
    \item[$\Box$] The debate is logically clear but weak in value expression.
    \item[$\Box$] The debate is value-rich but not well organized.
    \item[$\Box$] The debate does not clearly express or organize values.
\end{itemize}

\subsubsection*{Optional Comment}

Please briefly explain your evaluation, especially if you gave very high or very low scores.

\section{The Use of Large Language Models}

We used AI writing assistants (Claude and ChatGPT) for language polishing and improving the clarity of the manuscript. We also used GitHub Copilot to assist with implementation of the annotation pipeline and metric computation code. All scientific content, experimental design, analytical frameworks, results, and conclusions are entirely the work of the authors. AI assistants were not used to generate any experimental results, annotations, or analysis reported in this paper. 

\vspace{1.5cm}
\end{document}